\pdfoutput=1
\PassOptionsToPackage{table,xcdraw}{xcolor}
\documentclass[11pt]{article}

\usepackage[preprint]{acl}

\usepackage{times}
\usepackage{latexsym}

\usepackage[T1]{fontenc}

\usepackage[utf8]{inputenc}

\usepackage{microtype}

\usepackage{graphicx}

\usepackage{multirow} 
\usepackage{enumitem} 
\setlist[itemize]{noitemsep, topsep=0pt}
\usepackage{marvosym}
\usepackage[table,xcdraw]{xcolor}
\usepackage[normalem]{ulem} 
\useunder{\uline}{\ul}{} 
\usepackage{comment}
\usepackage{amssymb}
\usepackage{amsmath}
\usepackage{pifont}
\newcommand\pmcraw{$\text{PMC}^{Raw}$}
\newcommand\pmcrefine{$\text{PMC}^{Refined}$}
\newcommand\recipe{Recipe1M}
\newcommand\ours{{AdaMLLM}}
\definecolor{lightgray}{HTML}{EFEFEF} 
\usepackage{bm}
\usepackage{textcomp}
\makeatletter
\def\thanks#1{\protected@xdef\@thanks{\@thanks
        \protect\footnotetext{#1}}}
\makeatother

\usepackage{stfloats}
\usepackage{float} 
\usepackage{booktabs}
\usepackage{diagbox}
\usepackage{textgreek}

\title{On Domain-Adaptive Post-Training for Multimodal \\ Large Language Models}

\author{Daixuan Cheng\textsuperscript{\textalpha} 
~~
Shaohan Huang\textsuperscript{\textbeta}
~~
Ziyu Zhu\textsuperscript{\textgamma}
~~
Xintong Zhang\textsuperscript{\textepsilon}
\\[3pt]
\textbf{Wayne Xin Zhao\textsuperscript{\texttheta}
~~
Zhongzhi Luan\textsuperscript{\textbeta} 
~~
Bo Dai\textsuperscript{\textalpha\dag}\thanks{ \textsuperscript{\dag} Corresponding Author.}
~~~
Zhenliang Zhang\textsuperscript{\textalpha\dag}} 
\vspace{2mm}\\
\textsuperscript{\textalpha}BIGAI\quad\textsuperscript{\textbeta}BUAA\quad\textsuperscript{\textgamma}THU\quad\textsuperscript{\textepsilon}BIT\quad \textsuperscript{\texttheta}RUC\\
\url{https://huggingface.co/AdaptLLM}
}

\begin{document}
\maketitle
\begin{abstract}
Adapting general multimodal large language models (MLLMs) to specific domains, such as scientific and industrial fields, is highly significant in promoting their practical applications. This paper systematically investigates \textit{{domain adaptation of MLLMs via post-training}}, focusing on data synthesis, training pipeline, and task evaluation. (1) \textbf{Data Synthesis}: Using only open-source models, we develop a generate-then-filter pipeline that curates diverse visual instruction tasks based on domain-specific image-caption pairs. The resulting data surpass the data synthesized by manual rules or strong closed-source models in enhancing domain-specific performance. (2) \textbf{Training Pipeline}: Unlike general MLLMs that typically adopt a two-stage training paradigm, we find that a single-stage approach is more effective for domain adaptation. (3) \textbf{Task Evaluation}: We conduct extensive experiments in high-impact domains such as biomedicine, food, and remote sensing, by post-training a variety of MLLMs and then evaluating MLLM performance on various domain-specific tasks. Finally, we fully open-source our models, code, and data to encourage future research in this area.
\end{abstract}

\section{Introduction}
\label{sec:intro}
\begin{figure}[!htb]
    \centering
    \includegraphics[width=\linewidth]{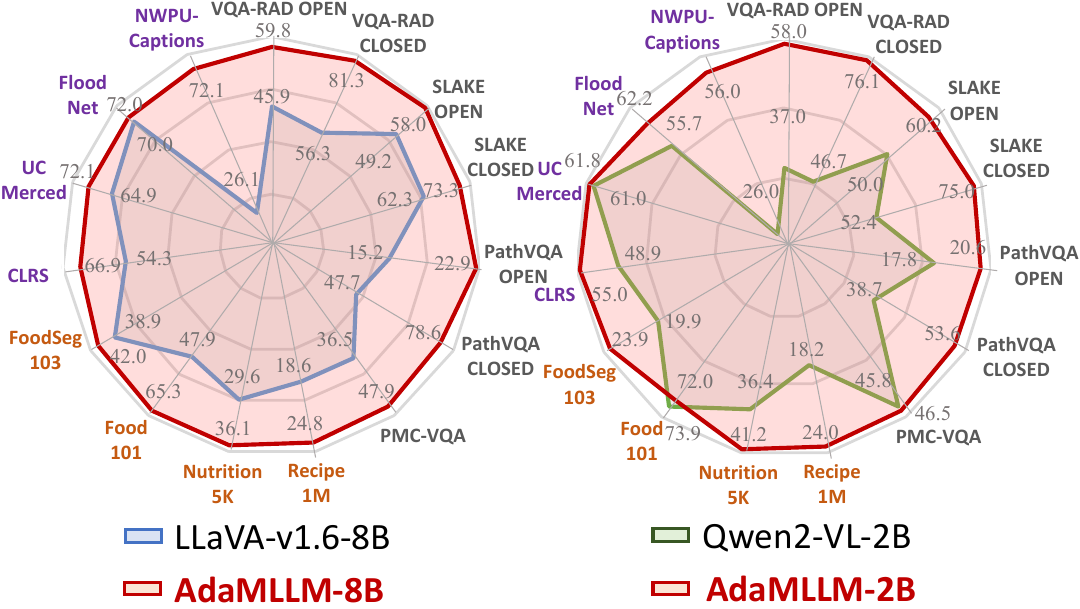}
    \vspace{-16pt}
    \caption{\textbf{Domain-Specific Task Performance of General MLLM and \ours.} For each domain, we conduct post-training to adapt the general MLLM and evaluate MLLM performance on various domain-specific tasks. {\color[HTML]{656565}Biomedicine}, {\color[HTML]{CE6301}food} and {\color[HTML]{800080}remote sensing} tasks are colored {\color[HTML]{656565}gray}, {\color[HTML]{CE6301}orange} and {\color[HTML]{800080}purple}, respectively.} 
\label{fig:intro}
\vspace{-15pt}
\end{figure}

General multimodal large language models (MLLMs;~\citealp{flamingo,llava}) have shown impressive capabilities in general scenarios. However, their expertise plummets in specialized domains due to insufficient domain-specific training~\citep{adaptllm}. For instance, scientific fields require learning from specialized images not commonly found in general scenarios~\citep{biomedgpt}; and industrial applications face privacy constraints that limit data access for general training~\citep{fintral}.

Domain-specific training for MLLMs requires diverse visual instruction tasks infused with domain knowledge~\citep{vilt-survey}. For domain-specific instruction synthesis, while using closed- or open-source models to synthesize data is common in training general MLLMs, challenges arise when applying this approach to domain-specific MLLMs due to privacy concerns with closed-source models~\citep{gpt-4v} and the lack of domain expertise in open-source models. For domain-specific training, the two-stage pipeline—first training on image-caption pairs, then on visual instruction tasks—is widely adopted for developing general MLLMs~\citep{llava}. However, tasks in specialized domains are often limited, and splitting them into two stages can further reduce task diversity within each stage.
 
In this paper, we systematically investigate domain-specific instruction synthesis and training pipelines for MLLM post-training, referred to as \textit{domain-adaptive training}. Specifically, we aim to (1) leverage only open-source models for data synthesis to avoid the privacy concerns of closed-source models and (2) maintain domain-specific task diversity throughout the post-training stage. 

\textbf{For instruction synthesis} with open-source models, we propose a generate-then-filter pipeline to address the lack of domain expertise. First, an open-source MLLM is fine-tuned to output instruction-response pairs based on input domain-specific image-caption pairs. To ensure diversity, we curate a seed data collection encompassing various domains and tasks based on existing datasets, without the need for additional expert annotation. The fine-tuned MLLM can effectively leverage domain knowledge in the image-caption source to generate diverse instruction-response pairs. Then, to ensure synthetic response accuracy, instead of directly verifying each response based on the instruction—which requires significant expertise—we propose selecting tasks with inherently consistent responses. This significantly improves accuracy while reducing the need for expert annotation. Although generated from open-source models, our synthetic tasks improve model performance more effectively than those generated by manual rules and strong closed-source models. 
\textbf{For training pipeline}, in the area of \textit{domain-adaptive post-training} of MLLMs, two-stage training remains mainstream~\citep{llava-med,pubmedvision,llava-chef}. However, we find that splitting the training data into two separate stages may hinder training task diversity and efficiency. Therefore, we apply a single-stage training pipeline that combines the synthetic task with its corresponding image-caption pair. This simple method enriches task diversity during training and leads to better performance.

In contrast to previous works~\citep{medflamigo, pmc-vqa, dolphins} which focus on a single domain or a single series of MLLMs per work, we conduct experiments across a variety of high-impact domains, such as biomedicine, food, and remote sensing, on general MLLMs of different sources and scales, such as Qwen2-VL-2B~\citep{qwen2-vl}, LLaVA-v1.6-8B~\citep{llava-v1.6}, and Llama-3.2-VL-11B~\citep{llama3}. As shown in Figure~\ref{fig:intro}, our resulting model, \ours~(short for \textbf{Ada}pted \textbf{M}ultimodal \textbf{L}arge \textbf{L}anguage \textbf{M}odel), consistently outperforms general MLLMs across various domain-specific tasks.

In summary, our contributions include:
\begin{itemize}[leftmargin=*]
\itemsep0em 
\item To the best of our knowledge, we present the first systematic investigation of \textit{domain-adaptive post-training} of MLLMs through extensive experiments across diverse domains and MLLMs.

\item We comprehensively analyze the data synthesis pipeline and training strategy, revealing that task diversity and domain knowledge are key to the success of our method.

\item We fully open-source our models, code, and data to facilitate future research and easy adaptation to new MLLMs and domains.  
\end{itemize}

\section{Related Work}
Our work is related to domain-specific MLLM, instruction synthesis, and MLLM training strategy. This section focuses on domain-specific MLLM, and the other topics are discussed in Appendix~\ref{app:Extended Related Work}.

\paragraph{Domain-Specific Data}
Initially, \citet{medflamigo} utilized multimodal interleaved data. With the rise of visual instruction tuning, research has shifted to synthesizing visual instructions~\citep{vilt-survey}. The approaches fall into two categories: (1) transforming existing datasets into visual instruction formats~\citep{llava-chef,foodlmm} (2) prompting closed-source models to generate tasks from images/annotations~\citep{pmc-vqa,llava-med,pubmedvision}. Our work aligns with the second in synthesizing domain-specific data based on image-caption pairs, but we utilize open-source models to avoid privacy concerns and lead to even better performance.

\paragraph{Domain-Specific Training}
One type of domain-specific training begins with an unaligned LLM and visual encoder~\citep{pmc-vqa}. Another type is post-training which starts with a well-aligned general MLLM~\citep{llava-med}. Compared to the first type, post-training is more efficient in terms of data and computation, making it our preferred method. In domain-specific post-training, previous works~\citep{llava-med,llava-chef,pubmedvision} adopt the two-stage training pipeline originally proposed for general MLLMs. We simplify this into a single stage to enhance task diversity within the training phase.

\section{Method}
\begin{figure*}[!htb]
    \centering
    \includegraphics[width=\linewidth]{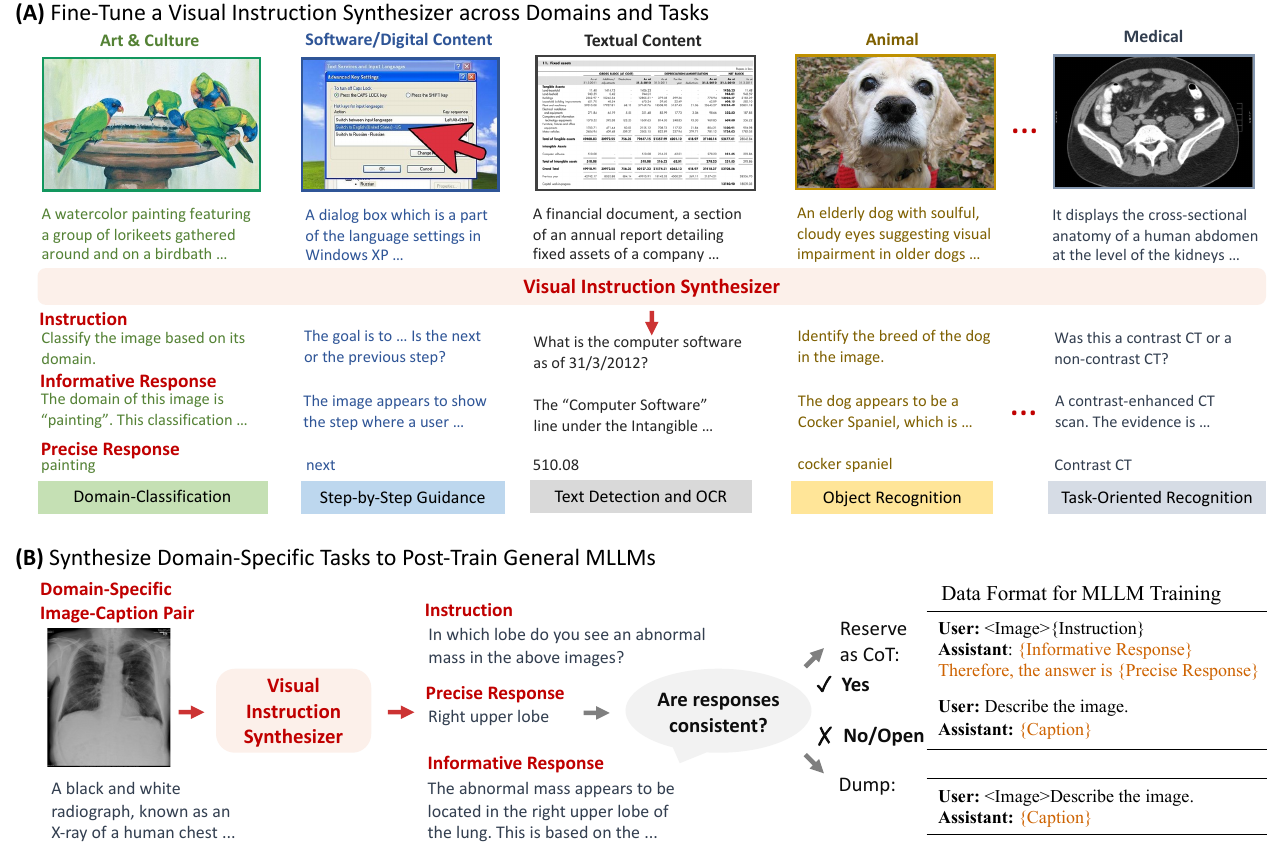}
    \caption{\textbf{Method Overview.} (A) We fine-tune a unified visual instruction synthesizer that generates diverse tasks based on image-caption pairs across various domains. (B) Using this synthesizer, we synthesize tasks based on domain-specific image-caption pairs and then apply a consistency-based data filter. The filtered synthetic tasks, combined with the original image captioning tasks, are employed to train general MLLMs through a single-stage post-training process, MLLM {\color[HTML]{CE6301}training loss} is computed only on the part colored in {\color[HTML]{CE6301}orange}.
}
\label{fig:method}
\vspace{-10pt}
\end{figure*}

We adapt MLLMs to domains via post-training. Figure~\ref{fig:method} provides the method overview: we begin by synthesizing domain-specific tasks using a unified visual instruction synthesizer, followed by a consistency-based data filter. The synthetic tasks are then combined with image-captioning tasks into a single stage for post-training.

\subsection{Domain Visual Instruction Synthesis}
The effectiveness of visual instruction tasks depends on diversity and accuracy, with domain knowledge being essential for domain adaptation~\citep{vilt-survey}. To meet these requirements, we propose a data synthesis approach comprising two components: a synthesizer that generates diverse tasks infused with domain knowledge, and a consistency-based filter to enhance accuracy.

\subsubsection{Visual Instruction Synthesizer} 
We fine-tune an open-source MLLM to generate diverse tasks based on image-caption pairs across various domains, developing a \textit{visual instruction synthesizer}. Instead of generating domain-specific tasks from scratch, which requires significant expertise, our synthesizer extracts tasks from existing data~\citep{instructpt}, thus reducing the reliance on domain experts. Furthermore, we incorporate specific designs to balance the utilization of image and text modalities.

\begin{table}[!tb]
\centering
\resizebox{\linewidth}{!}{\begin{tabular}{l}
\toprule
\begin{tabular}[c]{@{}l@{}}\textbf{User:} \textless{}Image\textgreater{}Describe the image.\\ \textbf{Assistant:} \{Caption\}\\\vspace{-8pt}\\ \color[HTML]{CE6301}\textbf{User:} Answer with a precise response. \{Instruction\}\\ \color[HTML]{CE6301}\textbf{Assistant:} \{Precise Response\}\\\vspace{-8pt}\\ \color[HTML]{CE6301}\textbf{User:} Answer with an informative response. \{Instruction\}\\\color[HTML]{CE6301} \textbf{Assistant:} \{Informative Response\}\end{tabular} \\ \bottomrule
\end{tabular}
}
\caption{\textbf{Data Format for Synthesizer Tuning.} The prefixes ``User'' and ``Assistant'' are determined by synthesizer's chat template. {\color[HTML]{CE6301}Tuning loss} is computed only on the part colored in {\color[HTML]{CE6301}orange}.}
\vspace{-15pt}
\label{tab:tune_syn}
\end{table}

\paragraph{Seed Data across Domains and Tasks}
We convert existing datasets~\citep{vision-flan,allava} into the seed data for fine-tuning our synthesizer, {\ul without any additional annotation}. As shown in part (A) of Figure~\ref{fig:method}, each seed data example includes an image-caption pair as input and a related task triplet---comprising an instruction, an informative response, and a precise response---as output. Compared to the precise response which is often a single phrase, the informative response contains more reasoning steps. The seed data cover a wide range of $20$ image domains and $191$ tasks. Details on data construction are in Appendix~\ref{app:Seed Data}.

\paragraph{Modality-Balanced Multitask Tuning}  
We fine-tune an open-source MLLM on the seed data to generate task triplets from image-caption pairs. As shown in Table~\ref{tab:tune_syn}, each seed data example is converted into a multi-turn conversation to fit the MLLM’s conversational format. During fine-tuning, we calculate the tuning loss only on the conversational turns related to the task triplet, ensuring the synthesizer focuses on them.

Furthermore, since the task instruction annotations in the seed data~\citep{vision-flan} rely solely on images—biasing the fine-tuned synthesizer toward over-reliance on visual inputs—we replace 10\% of the fine-tuning images with blank ones. This modality-balancing strategy encourages the model to leverage textual captions when visual inputs are ambiguous or uninformative, while preserving the quality of synthetic tasks for both complete and text-corrupted inputs. A detailed analysis is provided in Appendix~\ref{app:Further Ablations in Replacing Images with a Blank One}.

\paragraph{Task Synthesis for Target Domain}
After tuning, we use the synthesizer to generate task triplets from image-pairs in the target domain. For each image-caption pair, we input it into the synthesizer using the conversational format in Table~\ref{tab:tune_syn} and extract the task triplet from the output accordingly.

\subsubsection{Consistency-Based Filter}
Developed from an open-source model without sufficient domain expertise, our synthesizer inevitably produces some inaccurate responses, necessitating data filtering. We propose a filtering method based on {inherent consistency}, which improves data quality while reducing the need for expert validation.

As shown in part (B) in Figure~\ref{fig:method}, we prompt an open-source language model to classify each task triplet into one of three categories: consistent, inconsistent, or open. The consistent and inconsistent categories indicate whether the precise and informative responses align, while the open category indicates tasks that request open-ended responses (e.g., background information). The prompt template is in Figure~\ref{fig:consistency_prompt_1} in Appendix. We discard triplets classified as inconsistent, and those classified as open due to their ambiguity. In contrast to ensemble methods (multi-model voting;~\citealp{ensemble}) and self-consistency (multi-output sampling;~\citealp{Self-consistency}), we select each task triplet based on a single output from a single synthesizer, leveraging internal consistency within that output.

For consistent triplets, we combine the informative and precise responses into a chain-of-thought (CoT;~\citealp{cot}) format. The informative response serves as the reasoning process, and the precise response serves as the final conclusion, ensuring both informativeness and accuracy.

\subsection{Single-Stage Post-Training}
Domain-specific post-training for MLLMs~\citep{llava-med,pubmedvision,llava-chef} typically follows the multi-stage paradigm used in general MLLM training: first on image-caption pairs, then on visual instruction tasks~\citep{llava}. However, task diversity in domain-specific training is often more limited than in general training, and splitting the training into two stages may further reduce diversity within each stage, negatively impacting the task generalization of the trained models~\citep{flan}.  To mitigate this, we propose combining the training data into a single stage. As shown in part (B) of Figure~\ref{fig:method}, each training example includes two tasks:
\begin{itemize}[leftmargin=*]
\item \textit{Image Captioning Task}: A question prompting the MLLM to describe the image is randomly chosen from a pool in~\citep{allava} as the task instruction, with the original caption as the ground-truth response.
\item \textit{Synthetic Visual Instruction Task}: For each image-caption pair with a synthetic task after filtering, we combine it with the image captioning task in a multi-turn format. If no synthetic task remains, only the captioning task is used.
\end{itemize}

Following~\cite{llava}, we train on the data using the next-token prediction objective~\citep{gpt1}, computing loss only on the response part of each instruction-response pair.

\begin{table*}[h]
\centering
\resizebox{0.9\linewidth}{!}{\begin{tabular}{lccccccc}
\toprule
\multirow{2}{*}{\textbf{Biomedicine}} & \multicolumn{2}{c}{\small\textbf{SLAKE}} & \multicolumn{2}{c}{\small\textbf{PathVQA}} & \multicolumn{2}{c}{\small\textbf{VQA-RAD}} & \multirow{2}{*}{\small\textbf{PMC-VQA}} \\ \cmidrule(lr){2-3} \cmidrule(lr){4-5} \cmidrule(lr){6-7}
                                 & OPEN         & CLOSED        & OPEN         & CLOSED         & OPEN         & CLOSED                   &                                   \\ \midrule
\cmidrule{1-8}
\textbf{\textit{LLaVA-v1.6-8B}}         & 49.2       & 62.3         & 15.2        & 47.7      & 45.9        & 56.3      & 36.5                     \\
~~LLaVA-Med-8B                          & 43.4       & 50.2         & 10.1        & 59.2       & 35.0        & 62.5    & 37.1                     \\
~~PubMedVision-8B                       & 50.0       & 68.3         & 17.0        & 67.5       & 43.3        & 67.3    & 40.4                     \\
\rowcolor[HTML]{EFEFEF} 
~~AdaMLLM-8B~from \pmcraw                &{\ul56.8}      & \textbf{76.4}         & {\ul19.7}       & \textbf{79.3}     & {\ul51.0}        & {\ul80.5}       & {\ul 44.3}                    \\
\rowcolor[HTML]{EFEFEF} 
~~AdaMLLM-8B~from \pmcrefine              & \textbf{58.0}       & {\ul73.3}         & \textbf{22.9}        & {\ul78.6}      & \textbf{59.8}        & \textbf{81.3}      & \textbf{47.9}                     \\ \cmidrule{1-8}
\textbf{\textit{Qwen2-VL-2B}}                & 50.0       & 52.4         & 17.8        & 38.7       & 37.0        & 46.7      & {\ul45.8}                     \\
~~LLaVA-Med-2B                          & 43.4       & 55.5         & 11.8        & 60.1     & 37.1        & 58.8        & 41.2                     \\
~~PubMedVision-2B                         & 45.2       & 63.2         & 18.2        & \textbf{64.7}     & 41.3        & 67.3      & 43.2                     \\
\rowcolor[HTML]{EFEFEF} 
~~AdaMLLM-2B~from \pmcraw                 & {\ul53.2}       & \textbf{75.2}         & {\ul20.1}        & {\ul63.8}   & {\ul49.8}        & {\ul74.6}         & 43.5                     \\
\rowcolor[HTML]{EFEFEF} 
~~AdaMLLM-2B~from \pmcrefine             & \textbf{60.2}       & {\ul75.0}         & \textbf{20.6}        & 53.6       & \textbf{58.0}        & \textbf{76.1}      & \textbf{46.5}                     \\\cmidrule{1-8}
\textbf{\textit{Llama-3.2-11B}}         & 56.2       & 63.9         & {\ul22.7}        & 72.1      & 46.9        & 63.6      & \textbf{51.9}                     \\
~~LLaVA-Med-11B                            & 47.6       & 58.7         & 14.6        & 69.5      & 38.0        & 69.1     & 47.5                     \\
~~PubMedVision-11B                                                    & 49.1                   & 74.3                     & 19.3                    & 70.9             & 46.2                    & 73.9              & 47.1                                              \\
\rowcolor[HTML]{EFEFEF} 
~~AdaMLLM-11B from \pmcraw                & {\ul56.7}       & \textbf{77.6}         & 22.2        & \textbf{87.3}      & {\ul55.0}        & {\ul76.1}       & {\ul49.9}                     \\
\rowcolor[HTML]{EFEFEF} 
~~AdaMLLM-11B from \pmcrefine              & \textbf{59.5}       & {\ul76.4}         & \textbf{24.3}        & {\ul84.9}      & \textbf{57.4}        & \textbf{79.8}      & \textbf{51.9}                    \\\bottomrule
\end{tabular}
}
\caption{\textbf{Biomedicine Task Performance} of general MLLMs and MLLMs after domain-adaptive post-training. The image-caption sources for \colorbox{lightgray}{AdaMLLM from~\pmcraw}~and \colorbox{lightgray}{AdaMLLM from~\pmcrefine} are \pmcraw~and \pmcrefine, respectively.}
\label{tab:main_results_med}
\vspace{-10pt}
\end{table*}

\section{Experiment Settings}
We conduct experiments in high-impact domains including biomedicine, food, and remote sensing. For each domain, we perform post-training to adapt general MLLMs and evaluate model performance on various domain-specific tasks.

Note that our implementation is fully open-source, and the only required change for applying our method to a new domain is to replace the image-caption source with that of the new domain.

\paragraph{Image-Caption Data Source}  
For biomedicine domain, we use two sources: \pmcraw~in~\citet{llava-med} and \pmcrefine~in~\citet{pubmedvision}. For food domain, we collect data from \recipe~\citep{recipe1m}. For remote sensing domain, we collect data from five image-captioning tasks. Data details are in Appendix~\ref{app:Image-Caption Data Source}.

\paragraph{Visual Instruction Synthesis}
Our synthesizer is fine-tuned from {LLaVA-v1.6-Llama3-8B}. For the consistency-based filter, we prompt Llama-3-8B~\citep{llama3} to evaluate the consistency of each synthesized task triplet. Detailed implementations and costs are in Appendix~\ref{app:Visual Instruction Synthesizer}.

\paragraph{Post-Training \& Task Evaluation}  
Using synthetic data from the {LLaVA-v1.6-Llama3-8B}-based synthesizer, we conduct post-training on {LLaVA-v1.6-Llama3-8B} itself. Besides, we use the same synthetic data to post-train {Qwen2-VL-2B-Instruct} and {Llama-3.2-11B-Vision-Instruct} to assess effectiveness across different models and scales. Training hyper-parameters and costs are in Appendix~\ref{app:MLLM Post-Training Settings}. 

After post-training, we evaluate MLLMs on domain-specific tasks without further fine-tuning. Evaluation details are in Appendix~\ref{app:Task Evaluation Details}.

\paragraph{Baseline}
For biomedicine domain, we compare with two baselines: (1) {LLaVA-Med}~\citep{llava-med} which uses text-only GPT-4 to synthesize tasks from \pmcraw, and (2) {PubMedVision}~\citep{pubmedvision} which uses GPT-4V to synthesize tasks from \pmcrefine. For food domain, we compare with {LLaVA-Chef}~\citep{llava-chef} which uses manual rules to transform image-recipe pairs from~\recipe~into multiple tasks. LLaVA-Med, PubMedVision and LLaVA-Chef all employ two-stage post-training. For remote sensing, we compare with a baseline that uses GPT-4o~\citep{gpt-4o} to synthesize instructions based on the same image-caption pairs as ours, and train MLLMs using the same single-stage post-training. The resulting model is referred to as RS-4o.

\begin{table}[!t]
\centering
\resizebox{\columnwidth}{!}{\begin{tabular}{lcccc}
\toprule
\textbf{Food}                 & {\small\textbf{Recipe}} & {\small\textbf{Nutrition}} & {\small\textbf{Food101}} & {\small\textbf{FoodSeg}} \\\midrule
\textbf{\textit{LLaVA-v1.6-8B}}          & 18.6     & {\ul29.6}         & {\ul47.9}    & {\ul38.9}       \\
~~LLaVA-Chef-8B                  & {\ul23.1}     & 29.1         & 46.8    & 14.5       \\
\rowcolor[HTML]{EFEFEF} 
~~AdaMLLM-8B                        & \textbf{24.8}     & \textbf{36.1}         & \textbf{65.3}    & \textbf{42.0}       \\\cmidrule{1-5}
\textbf{\textit{Qwen2-VL-2B}}         & 18.2     & {\ul36.4}         & \textbf{73.9}    & {\ul19.9}       \\
~~LLaVA-Chef-2B                  & \textbf{24.1}     & 24.5         & 68.8    & 7.7        \\
\rowcolor[HTML]{EFEFEF} 
~~AdaMLLM-2B                        & {\ul24.0}     & \textbf{41.2}         & {\ul72.0}    & \textbf{23.9}       \\\cmidrule{1-5}
\textbf{\textit{Llama-3.2-11B}} & 23.7     & {\ul40.0}         & 80.8    & \textbf{47.6}       \\
~~LLaVA-Chef-11B                  & {\ul25.7}     & 26.2         & {\ul82.1}    & 16.7       \\
\rowcolor[HTML]{EFEFEF} 
~~AdaMLLM-11B                      & \textbf{26.1}     & \textbf{41.0}         & \textbf{82.2}    & {\ul42.0}    \\  \bottomrule
\end{tabular}
}
\vspace{10pt}

\resizebox{\columnwidth}{!}{\begin{tabular}{lcccc}
\toprule
\textbf{Remote Sensing}                 & {\small\textbf{CLRS}} & {\small\textbf{UC Merced}} & {\small\textbf{FloodNet}} & {\small\textbf{NWPU}} \\\midrule
\textbf{\textit{LLaVA-v1.6-8B}}          & {\ul54.3}     & {\ul 64.9}         & {\ul70.0}    & {26.1}    \\
~~RS-4o-8B                  & {50.3}     & 64.5        & 58.1     & \textbf{74.2}       \\
\rowcolor[HTML]{EFEFEF} 
~~AdaMLLM-8B                        & \textbf{66.9}     & \textbf{72.1}         & \textbf{72.0 }    & {\ul72.1}      \\\cmidrule{1-5}
\textbf{\textit{Qwen2-VL-2B}}         & 48.9     & {61.0}         & {\ul55.7}    & 26.0        \\
~~RS-4o-2B                  & {\ul51.2}     & \textbf{67.0}         & 53.7    & \textbf{56.7}        \\
\rowcolor[HTML]{EFEFEF} 
~~AdaMLLM-2B                        & \textbf{55.0}     & {\ul61.8}         & \textbf{62.2}    & {\ul56.0}         \\\cmidrule{1-5}
\textbf{\textit{Llama-3.2-11B}} & 55.7      & {\ul74.2}          & {\ul60.8}     & 20.8       \\
~~RS-4o-11B                  & {\ul59.3}      & 57.6          & 51.5     & \textbf{71.9}      \\
\rowcolor[HTML]{EFEFEF} 
~~AdaMLLM-11B                      & \textbf{64.9}     & \textbf{81.5}         & \textbf{62.1}    & {\ul67.8}    \\  \bottomrule
\end{tabular}

}
\caption{\textbf{Food and Remote Sensing Task Performance} of general MLLMs and MLLMs after domain-adaptive post-training.}
\label{tab:main_results_food}
\vspace{-15pt}
\end{table}

\begin{table*}[h]
\centering
\resizebox{0.9\linewidth}{!}{\begin{tabular}{lcccccccccccc}
\toprule
\textbf{Image-Caption} & \multicolumn{4}{c}{\textbf{\recipe}}                                                 & \multicolumn{4}{c}{\textbf{\pmcraw}}                                          & \multicolumn{4}{c}{\textbf{\pmcrefine}}                                       \\ \cmidrule(lr){2-5} \cmidrule(lr){6-9} \cmidrule(lr){10-13}
\textit{Train Pipeline}    & \multicolumn{2}{c}{\textit{Two-stage}}    & \multicolumn{2}{c}{\colorbox{lightgray}{\textit{Single-stage}}}    & \multicolumn{2}{c}{\textit{Two-stage}} & \multicolumn{2}{c}{\colorbox{lightgray}{\textit{Single-stage}}} & \multicolumn{2}{c}{\textit{Two-stage}} & \multicolumn{2}{c}{\colorbox{lightgray}{\textit{Single-stage}}} \\\cmidrule(lr){2-3} \cmidrule(lr){4-5} \cmidrule(lr){6-7} \cmidrule(lr){8-9} \cmidrule(lr){10-11} \cmidrule(lr){12-13}
Instruction   & Rule & \cellcolor[HTML]{EFEFEF}Ours & Rule & \cellcolor[HTML]{EFEFEF}Ours & GPT-4   & \cellcolor[HTML]{EFEFEF}Ours     & GPT-4    & \cellcolor[HTML]{EFEFEF}Ours     & GPT-4V     & \cellcolor[HTML]{EFEFEF}Ours    & GPT-4V     & \cellcolor[HTML]{EFEFEF}Ours    \\ \midrule
LLaVA-v1.6-8B                 & 28.4                      & \cellcolor[HTML]{EFEFEF}\textbf{29.0} & 34.1                      & \cellcolor[HTML]{EFEFEF}\textbf{42.0}~\textuparrow & 42.5               & \cellcolor[HTML]{EFEFEF}\textbf{55.6}     & 46.1               & \cellcolor[HTML]{EFEFEF}\textbf{58.3}~\textuparrow     & 50.5                & \cellcolor[HTML]{EFEFEF}\textbf{58.6}    & 55.5                & \cellcolor[HTML]{EFEFEF}\textbf{60.3}~\textuparrow    \\
Qwen2-VL-2B                  & 31.3                      & \cellcolor[HTML]{EFEFEF}\textbf{38.2} & 31.9                      & \cellcolor[HTML]{EFEFEF}\textbf{40.3}~\textuparrow & 44.0               & \cellcolor[HTML]{EFEFEF}\textbf{55.5}     & 41.3               & \cellcolor[HTML]{EFEFEF}\textbf{54.3}~\textdownarrow     & 49.0                & \cellcolor[HTML]{EFEFEF}\textbf{59.5}    & 51.6                & \cellcolor[HTML]{EFEFEF}\textbf{55.7}~\textdownarrow    \\
Llama-3.2-11B                 & 37.7                      & \cellcolor[HTML]{EFEFEF}\textbf{40.9} & 36.6                      & \cellcolor[HTML]{EFEFEF}\textbf{47.8}~\textuparrow & 49.3               & \cellcolor[HTML]{EFEFEF}\textbf{59.2}     & 48.8               & \cellcolor[HTML]{EFEFEF}\textbf{60.7}~\textuparrow     & 54.4                & \cellcolor[HTML]{EFEFEF}\textbf{60.3}    & 53.7                & \cellcolor[HTML]{EFEFEF}\textbf{62.0}~\textuparrow   \\ \bottomrule
\end{tabular}
}
\caption{\textbf{Domain-Specific Task Performance of MLLMs after Post-Training} with different synthetic data and training pipelines. We report the average performance in each domain, with detailed results in Table~\ref{tab:detailed Comparison of Synthetic Tasks and Training Pipeline} in Appendix. When the image-caption source and training pipeline are fixed, synthetic data of better performance are marked in \textbf{bold}. When the image-caption source is fixed and our synthetic data are used, numbers marked with \textuparrow~indicate that single-stage training outperforms two-stage training, while \textdownarrow~indicates the opposite.}
\label{tab:stage_method}
\vspace{-5pt}
\end{table*}

\begin{table*}[h]
\centering
\resizebox{\linewidth}{!}{\begin{tabular}{lccccccccc}
\toprule
         & \multirow{2}{*}{{\textbf{Ours}}}                               & \multirow{2}{*}{\textbf{{\begin{tabular}[c]{@{}c@{}}General \\ CoT Task \end{tabular}}}} & \multirow{2}{*}{\textbf{\begin{tabular}[c]{@{}c@{}}General CoT Task + \\ Domain Caption \end{tabular}}} & \multirow{2}{*}{\textbf{\begin{tabular}[c]{@{}c@{}}w/o\\ Blank Image\end{tabular}}} & \multicolumn{2}{c}{\textbf{w/o Consistency Filter}} & \multirow{2}{*}{\textbf{\begin{tabular}[c]{@{}c@{}} Image Caption \\ Only \end{tabular}}} & \multirow{2}{*}{\textbf{\begin{tabular}[c]{@{}c@{}}Synthetic Task \\ Only \end{tabular}}} & \multirow{2}{*}{\textbf{\begin{tabular}[c]{@{}c@{}}Two-Stage\\ Reuse Caption\end{tabular}}}  \\ \cmidrule(lr){6-7}  &                                              &                                            
        &                                &                                           & Precise              & Informative             &         &           \\ \midrule
BioMed. & \textbf{58.3} & 49.8             & 55.3                          & 55.8                                      & 31.2                       & 44.4                                              & 26.7                                         & 54.2                                 & 54.9                                 \\
Food    & \textbf{42.0}              & 36.0             & 38.6     & 35.9                                      & 37.9                       & 37.6                                            & 25.6                                       & 36.8     & 36.6   \\ \bottomrule                                    
\end{tabular}
}
\caption{\textbf{Ablation Results.} ``General CoT Task'' trains on seed data processed into our task format,``General CoT Task +  Domain Caption'' mixes the processed seed data with domain-specific image-caption pairs. ``w/o Blank Image'' fine-tunes the synthesizer without replacing 10\% of images with blank ones. ``w/o Consistency Filter'' removes the consistency-based filter and trains with either precise or informative responses. ``Image Caption Only'' removes synthetic task, and ``Synthetic Task Only'' removes image captioning task. ``Two-Stage Reuse Caption'' conducts two-stage training with the second stage reusing the caption data from the first stage. }
\label{tab:ablation}
\vspace{-5pt}
\end{table*}

\section{Main Results}
\label{sec:Main Results}

\paragraph{Overall Performance} As shown in Tables~\ref{tab:main_results_med} and~\ref{tab:main_results_food}, ours consistently enhances MLLM performance, outperforming baselines across various domain-specific tasks. Although our synthesizer is based on LLaVA-v1.6-8B, we observe consistent improvements on Qwen2-VL-2B and Llama-3.2-11B, demonstrating its effectiveness across different models and scales. Among the evaluated tasks, VQA-RAD, Recipe1M and NWPU can be regarded as partially seen tasks, with VQA-RAD included in our seed data for fine-tuning the synthesizer, Recipe1M and NWPU included in the image-caption source\footnote{Test/validation sets of VQA-RAD, Recipe1M and NWPU are not included.}. Nevertheless, \ours~shows consistent gains on other unseen tasks, demonstrating strong task generalization in the target domain.

\paragraph{Comparison of Synthetic Task and Training Pipeline}
In addition to the overall comparison, we assess the effectiveness of our synthetic tasks and single-stage training separately by varying one factor at a time. As shown in Table~\ref{tab:stage_method}, we conduct both two-stage and single-stage post-training with synthetic tasks generated by different methods: manual rules in {LLaVA-Chef}, GPT-4 in {LLaVA-Med}, and GPT-4V in {PubMedVision}. Our synthetic tasks consistently outperform others across both training pipelines. Furthermore, with our synthetic tasks, single-stage training surpasses two-stage training in most of the experiments.

\section{Ablations}
\label{sec:ablations}
To evaluate the effectiveness of each component, we conduct ablations to post-train LLaVA-v1.6-8B with different settings. We report the average task performance within each domain for the trained models in Table~\ref{tab:ablation}. 

\paragraph{Domain Knowledge, Task (CoT) Format \& Seed Data}
Our synthetic tasks are designed to integrate both (1) domain knowledge and (2) a visual instruction task format with Chain-of-Thought response. {To assess the role of domain knowledge}, we compare our method with general tasks that preserve the same CoT response format but exclude domain-specific knowledge. As shown in Table~\ref{tab:ablation}, our method outperforms the ``General CoT Task'', underscoring the importance of domain knowledge. 

Importantly, ``General CoT Task'' is constructed from our seed data—originally used to fine-tune the synthesizer—and filtered using the same consistency-based method. This suggests that the performance gain of our method is not simply due to knowledge distillation from the seed data. 

Moreover, our method outperforms ``General CoT Task + Domain Caption'', indicating that naively combining domain-specific image-caption pairs with general tasks is insufficient. In contrast, our synthesis pipeline effectively transforms the domain knowledge embedded in image-caption pairs into a format that general MLLMs can learn from more effectively.

\paragraph{Visual Instruction Synthesis}
To balance modality-utilization, we replace some of the images with blank images during the fine-tuning of synthesizer. The effectiveness of this strategy on MLLM performance is demonstrated in ``w/o Blank Image''.

To improve response accuracy, we design a consistency-based filter. As shown in Table~\ref{tab:ablation}, removing this filter results in decreased model performance, regardless of whether the response contains only precise or informative content.

\paragraph{Single-Stage Post-Training}
Our motivation for combining data into a single stage is to enhance training task diversity. This efficacy is evident in the ablation results in Table~\ref{tab:ablation}, where removing either the synthetic task (``Image Caption Only'') or the image captioning task (``Synthetic Task Only'') degrades model performance, even when the caption data is reused in the second stage (``Two-Stage Reuse Caption''). 

\section{Analysis}
\label{sec:Analysis}
We conduct a detailed analysis on the synthesis pipeline and the synthesized data

\begin{table}[!tb]
\centering
\resizebox{\columnwidth}{!}{\begin{tabular}{lccccc}
\toprule
\textbf{Finetune Input}   & -    & \textbf{Image} &\textbf{Caption} & \multicolumn{2}{c}{\textbf{Image + Caption}}                  \\\cmidrule(lr){5-6}
\textit{Blank Image} & -    & -   & -   &  \ding{55} & \ding{51}\\\midrule
{Diversity}         & 52.5 & 68.0  & 75.2  & 81.0 & \textbf{85.5} \\
{Knowledge}           & 72.5 & 95.0  & 93.8  & {97.5} & \textbf{98.1}\\
{Complexity}          & 43.8 & 77.9  & 75.3 & 80.0 & \textbf{83.2} \\
{Accuracy}          & 63.8 & 60.0  & 65.6 & 66.3 & \textbf{71.3} \\ \bottomrule
\end{tabular}
}
\caption{\textbf{Quality of Synthetic Tasks by Different Visual Instruction Synthesizers.} Column 1 presents results from the MLLM without fine-tuning (i.e., the base LLaVA in our experiment settings). Columns 2-5 show results after fine-tuning the MLLM using our seed data to synthesize tasks based on different inputs. Besides, Column 5 replaces 10\% of the images with blank ones.}
\label{tab:synthesizer}
\end{table}

\begin{table}[!tb]
\centering
\resizebox{\columnwidth}{!}{\begin{tabular}{lccccc}
\toprule
        & \multicolumn{3}{c}{\textbf{w/o Filter}}       & \multicolumn{2}{c}{\textbf{w/ Filter}} \\ \cmidrule(lr){2-4} \cmidrule(lr){5-6}
        & Consist. & Precise Acc & Info. Acc & Consist.       & Acc       \\ \midrule
BioMed. & 30.3        & 64.3        & 61.0     & 92.2             & 75.1  \\
Food    & 35.7        & 77.2        & 75.5     & 97.1             & 84.3   \\ \bottomrule    
\end{tabular}
}
\caption{\textbf{Quality of Responses with/without Using Consistency-Based Filter}, assessed in terms of consistency between precise and informative responses (Consist.), accuracy of precise responses (Precise Acc), accuracy of informative responses (Info. Acc), and accuracy of combined responses (Acc).}
\label{tab:filter}
\vspace{-10pt}
\end{table}

\subsection{Domain Visual Instruction Synthesis} 
\label{sec:Domain-Specific Visual Instruction Synthesis}
\paragraph{Visual Instruction Synthesizer} We compare tasks generated by synthesizers with different designs using a validation set from our seed data. Specifically, we conduct human evaluation of data quality in the following aspects: {task diversity}, {domain knowledge utilization}, {task complexity} and {response accuracy} (detailed definition and scoring criteria are in Appendix~\ref{app:Scoring Criteria for Data Quality}).
\begin{table}[!tb]
\centering
\resizebox{\columnwidth}{!}{
\begin{tabular}{lcccccc}
\toprule
\textbf{Image-Caption} & \multicolumn{2}{c}{\textbf{\recipe}} & \multicolumn{2}{c}{\textbf{\pmcraw}} & \multicolumn{2}{c}{\textbf{\pmcrefine}} \\  \cmidrule(lr){2-3} \cmidrule(lr){4-5} \cmidrule(lr){6-7}
Instruction            & Rule             & \cellcolor[HTML]{EFEFEF}Ours             & GPT-4              & \cellcolor[HTML]{EFEFEF}Ours              & GPT-4V               & \cellcolor[HTML]{EFEFEF}Ours               \\ \midrule
Diversity         & 23.5               & \cellcolor[HTML]{EFEFEF}\textbf{52.9}             & 47.1               & \cellcolor[HTML]{EFEFEF}\textbf{58.8}              & 64.7                 & \cellcolor[HTML]{EFEFEF}\textbf{76.5}               \\
Knowledge      & 20.9               & \cellcolor[HTML]{EFEFEF}\textbf{21.9}             & 44.9               & \cellcolor[HTML]{EFEFEF}\textbf{58.9}              & \textbf{67.7}                 & \cellcolor[HTML]{EFEFEF}{63.2}               \\
Complexity        & 38.4               & \cellcolor[HTML]{EFEFEF}\textbf{69.9}             & 41.7               & \cellcolor[HTML]{EFEFEF}\textbf{83.2}              & 49.6                 & \cellcolor[HTML]{EFEFEF}\textbf{80.5}               \\
Accuracy      & \textbf{98.7}               & \cellcolor[HTML]{EFEFEF}84.3             & \textbf{84.4}               & \cellcolor[HTML]{EFEFEF}75.1              & \textbf{87.5}                 & \cellcolor[HTML]{EFEFEF}79.6  \\ \bottomrule            
\end{tabular}
}
\caption{\textbf{Quality of Synthetic Tasks} by our method, manual rules, GPT-4, and GPT-4V.}
\label{tab:evaluation_data_quality}
\end{table}

\begin{figure}[!tb]
    \centering
    \includegraphics[width=\linewidth]{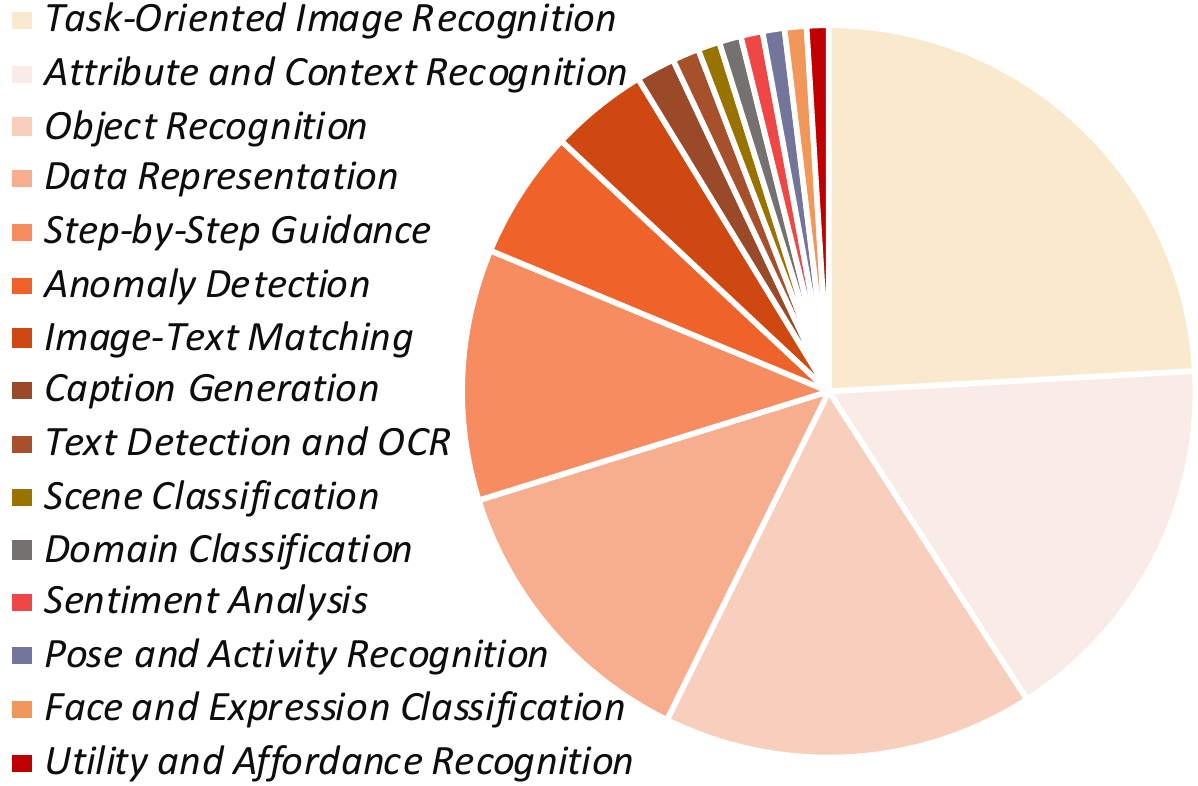}
    \vspace{-10pt}\caption{\textbf{Task Type Distribution} of all our synthetic tasks based on all image-caption sources.}
\label{fig:task_distribution}
\vspace{-10pt}
\end{figure}

\begin{figure*}[!htb]
    \centering
    \includegraphics[width=\linewidth]{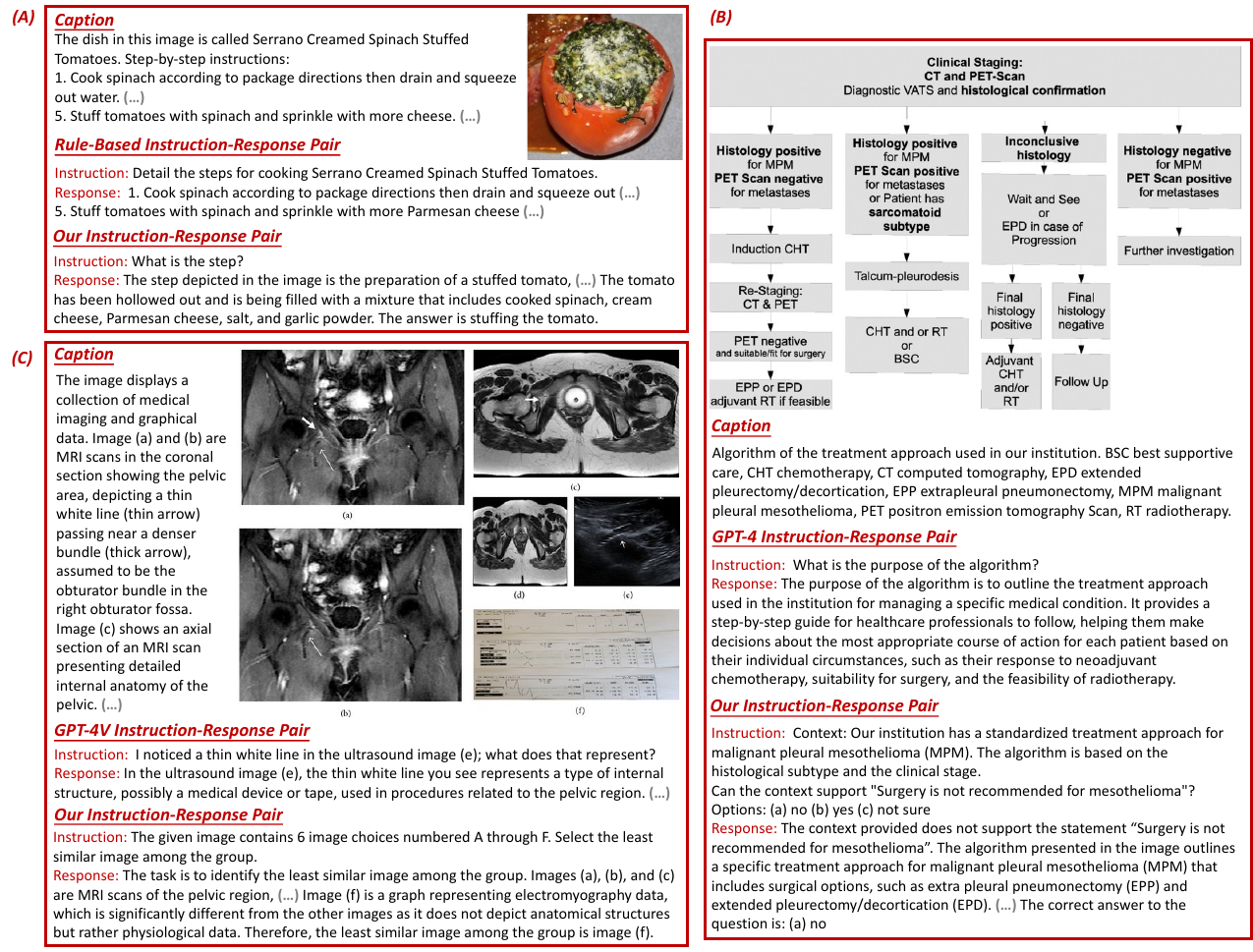}
    \vspace{-15pt}
    \caption{\textbf{Cases of Instruction-Response Pairs} synthesized by our method, manual rules, GPT-4, and GPT-4V.} 
\label{fig:case}
\vspace{-15pt}
\end{figure*}

The results in Table~\ref{tab:synthesizer} indicate that fine-tuning for task synthesis using either image \citep{genixer} or caption \citep{instructpt} inputs yields improvements in most aspects. Our design, which employs both image and caption inputs, leads to even higher performance. Besides, replacing 10\% of the images with blank ones achieves the highest quality (further analysis is in Appendix~\ref{app:Further Ablations in Replacing Images with a Blank One}).

\paragraph{Consistency-Based Filter}
Our consistency-based filter is designed to select tasks with inherent consistency, thereby increasing data accuracy. As shown in Table~\ref{tab:filter}, using the filter significantly increases the consistency between precise and informative responses, making the combination of them in the CoT format reasonable. As a result, the filter successfully increase response accuracy.

\subsection{Domain-Specific Synthetic Data}
\label{sec:Domain-Specific Synthetic Data}
\paragraph{Quantitative Analysis} 
Table~\ref{tab:evaluation_data_quality} presents the data quality scores for synthetic tasks generated by different methods. Our tasks are diverse and complex, demonstrating a high utilization of domain knowledge. The distribution of task types for all our instruction-response pairs is displayed in Figure~\ref{fig:task_distribution}. This explains the effectiveness of our method in enhancing MLLM performance across domain-specific tasks. However, our method underperforms the baselines in terms of response accuracy, with manual rules achieving nearly 100\% accuracy and GPT-4 and GPT-4V reaching around 85\%. This may because of the increased complexity of our synthesized tasks, which make generating accurate responses more challenging. These results indicate the need for further improvements to enhance response accuracy, even with complex tasks.

\paragraph{Qualitative Analysis} Figure~\ref{fig:case} presents cases of synthetic tasks by different methods when given the same image-caption pair. In case (A), the rule-based task is a simple transformation of the recipe caption, ignoring the image information. In contrast, our task conducts a detailed analysis of the food's state in the image and accurately matches it with the cooking step in the caption, demonstrating a higher level of {\ul domain knowledge utilization}. In case (B), both our task and the GPT-4 synthesized task focus on interpreting intent. While the GPT-4 task straightforwardly asks for the intent, our task increases {\ul task complexity} by requiring inference from the context to make a ``yes/no/not sure'' choice. In case (C) with multiple sub-images, our task type is distinct in requiring the identification of the least similar image among the group, showcasing {\ul task diversity}. More cases are provided in Figure~\ref{fig:more_case_1} in Appendix.

\section{Conclusion}
This paper investigates adapting general MLLMs to specific domains via post-training. To synthesize domain-specific visual instruction tasks, we develop a unified visual instruction synthesizer that generates instruction-response pairs based on domain-specific image-caption data, and then apply a consistency-based filter to improve data accuracy. This enables us to effectively synthesize diverse tasks with high domain knowledge utilization. For the post-training pipeline, we propose combining the synthetic tasks with image-captioning tasks into a single training stage to enhance task diversity. In multiple high-impact domains, our resulting model, AdaMLLM, consistently outperforms general MLLMs across various domain-specific tasks. 

\clearpage
\section*{Limitations}
While synthetic data reduce the need of expert annotation, it is crucial to acknowledge the potential limitations. Our work, along with other works utilizing synthetic data~\citep{syntheticsurvey}, is inevitably constrained by the possibility of introducing hallucinations. As shown in our analysis in Section~\ref{sec:Domain-Specific Synthetic Data}, the accuracy of our synthetic tasks remains imperfect, underscoring the need for further improvements to enhance response reliability, even with highly complex tasks.

Furthermore, future research may take into account the preferences of the specific domain. For instance, when dealing with animals, semantic-level information might be more important, while for medical images, local detail information should be given greater attention. 

\section*{Ethics Statement}
All the datasets and models used in this work are publicly available. 

\bibliography{ref}

\begin{thebibliography}{65}
\providecommand{\natexlab}[1]{#1}

\bibitem[{Alayrac et~al.(2022)Alayrac, Donahue, Luc, Miech, Barr, Hasson, Lenc,
  Mensch, Millican, Reynolds et~al.}]{flamingo}
Jean-Baptiste Alayrac, Jeff Donahue, Pauline Luc, Antoine Miech, Iain Barr,
  Yana Hasson, Karel Lenc, Arthur Mensch, Katherine Millican, Malcolm Reynolds,
  and 1 others. 2022.
\newblock Flamingo: a visual language model for few-shot learning.
\newblock \emph{Advances in neural information processing systems},
  35:23716--23736.

\bibitem[{Bhatia et~al.(2024)Bhatia, Nagoudi, Cavusoglu, and
  Abdul-Mageed}]{fintral}
Gagan Bhatia, El~Moatez~Billah Nagoudi, Hasan Cavusoglu, and Muhammad
  Abdul-Mageed. 2024.
\newblock Fintral: A family of gpt-4 level multimodal financial large language
  models.
\newblock \emph{arXiv preprint arXiv:2402.10986}.

\bibitem[{Bossard et~al.(2014)Bossard, Guillaumin, and Van~Gool}]{food101}
Lukas Bossard, Matthieu Guillaumin, and Luc Van~Gool. 2014.
\newblock Food-101--mining discriminative components with random forests.
\newblock In \emph{Computer vision--ECCV 2014: 13th European conference,
  zurich, Switzerland, September 6-12, 2014, proceedings, part VI 13}, pages
  446--461. Springer.

\bibitem[{Brown(2020)}]{gpt3}
Tom~B Brown. 2020.
\newblock Language models are few-shot learners.
\newblock \emph{arXiv preprint arXiv:2005.14165}.

\bibitem[{Chen et~al.(2024{\natexlab{a}})Chen, Chen, Zhang, Chen, Wu, Zhang,
  Chen, Li, Wan, and Wang}]{allava}
Guiming~Hardy Chen, Shunian Chen, Ruifei Zhang, Junying Chen, Xiangbo Wu, Zhiyi
  Zhang, Zhihong Chen, Jianquan Li, Xiang Wan, and Benyou Wang.
  2024{\natexlab{a}}.
\newblock Allava: Harnessing gpt4v-synthesized data for a lite vision-language
  model.
\newblock \emph{arXiv preprint arXiv:2402.11684}.

\bibitem[{Chen et~al.(2024{\natexlab{b}})Chen, Ouyang, Gao, Chen, Chen, Wang,
  Zhang, Cai, Ji, Yu et~al.}]{pubmedvision}
Junying Chen, Ruyi Ouyang, Anningzhe Gao, Shunian Chen, Guiming~Hardy Chen,
  Xidong Wang, Ruifei Zhang, Zhenyang Cai, Ke~Ji, Guangjun Yu, and 1 others.
  2024{\natexlab{b}}.
\newblock Huatuogpt-vision, towards injecting medical visual knowledge into
  multimodal llms at scale.
\newblock \emph{arXiv preprint arXiv:2406.19280}.

\bibitem[{Chen et~al.(2023)Chen, Li, Dong, Zhang, He, Wang, Zhao, and
  Lin}]{sharegpt4v}
Lin Chen, Jinsong Li, Xiaoyi Dong, Pan Zhang, Conghui He, Jiaqi Wang, Feng
  Zhao, and Dahua Lin. 2023.
\newblock Sharegpt4v: Improving large multi-modal models with better captions.
\newblock \emph{arXiv preprint arXiv:2311.12793}.

\bibitem[{Chen and Xing(2024)}]{open-llava-next}
Lin Chen and Long Xing. 2024.
\newblock \href {https://doi.org/10.5281/zenodo.13935471} {Open-llava-next: An
  open-source implementation of llava-next series for facilitating the large
  multi-modal model community.}
\newblock \url{https://github.com/xiaoachen98/Open-LLaVA-NeXT}.

\bibitem[{Cheng et~al.(2024{\natexlab{a}})Cheng, Gu, Huang, Bi, Huang, and
  Wei}]{instructpt}
Daixuan Cheng, Yuxian Gu, Shaohan Huang, Junyu Bi, Minlie Huang, and Furu Wei.
  2024{\natexlab{a}}.
\newblock Instruction pre-training: Language models are supervised multitask
  learners.
\newblock \emph{arXiv preprint arXiv:2406.14491}.

\bibitem[{Cheng et~al.(2024{\natexlab{b}})Cheng, Huang, and Wei}]{adaptllm}
Daixuan Cheng, Shaohan Huang, and Furu Wei. 2024{\natexlab{b}}.
\newblock \href {https://openreview.net/forum?id=y886UXPEZ0} {Adapting large
  language models via reading comprehension}.
\newblock In \emph{The Twelfth International Conference on Learning
  Representations}.

\bibitem[{Cheng et~al.(2022)Cheng, Huang, Xu, Zhou, Li, and Wang}]{NWPU}
Qimin Cheng, Haiyan Huang, Yuan Xu, Yuzhuo Zhou, Huanying Li, and Zhongyuan
  Wang. 2022.
\newblock Nwpu-captions dataset and mlca-net for remote sensing image
  captioning.
\newblock \emph{{IEEE} Trans. Geosci. Remote. Sens.}, 60:1--19.

\bibitem[{Chowdhery et~al.(2023)Chowdhery, Narang, Devlin, Bosma, Mishra,
  Roberts, Barham, Chung, Sutton, Gehrmann et~al.}]{palm}
Aakanksha Chowdhery, Sharan Narang, Jacob Devlin, Maarten Bosma, Gaurav Mishra,
  Adam Roberts, Paul Barham, Hyung~Won Chung, Charles Sutton, Sebastian
  Gehrmann, and 1 others. 2023.
\newblock Palm: Scaling language modeling with pathways.
\newblock \emph{Journal of Machine Learning Research}, 24(240):1--113.

\bibitem[{Dietterich(2000)}]{ensemble}
Thomas~G Dietterich. 2000.
\newblock Ensemble methods in machine learning.
\newblock In \emph{International workshop on multiple classifier systems},
  pages 1--15. Springer.

\bibitem[{Dubey et~al.(2024)Dubey, Jauhri, Pandey, Kadian, Al-Dahle, Letman,
  Mathur, Schelten, Yang, Fan et~al.}]{llama3}
Abhimanyu Dubey, Abhinav Jauhri, Abhinav Pandey, Abhishek Kadian, Ahmad
  Al-Dahle, Aiesha Letman, Akhil Mathur, Alan Schelten, Amy Yang, Angela Fan,
  and 1 others. 2024.
\newblock The llama 3 herd of models.
\newblock \emph{arXiv preprint arXiv:2407.21783}.

\bibitem[{Gu et~al.(2022)Gu, Ke, Zhu, and Huang}]{udit}
Yuxian Gu, Pei Ke, Xiaoyan Zhu, and Minlie Huang. 2022.
\newblock Learning instructions with unlabeled data for zero-shot cross-task
  generalization.
\newblock In \emph{Proceedings of the 2022 Conference on Empirical Methods in
  Natural Language Processing}, pages 1617--1634.

\bibitem[{He et~al.(2020)He, Zhang, Mou, Xing, and Xie}]{pathvqa}
Xuehai He, Yichen Zhang, Luntian Mou, Eric Xing, and Pengtao Xie. 2020.
\newblock Pathvqa: 30000+ questions for medical visual question answering.
\newblock \emph{arXiv preprint arXiv:2003.10286}.

\bibitem[{Huang et~al.(2023)Huang, Dong, Wang, Hao, Singhal, Ma, Lv, Cui,
  Mohammed, Patra et~al.}]{kosmos1}
Shaohan Huang, Li~Dong, Wenhui Wang, Yaru Hao, Saksham Singhal, Shuming Ma,
  Tengchao Lv, Lei Cui, Owais~Khan Mohammed, Barun Patra, and 1 others. 2023.
\newblock Language is not all you need: Aligning perception with language
  models.
\newblock \emph{Advances in Neural Information Processing Systems},
  36:72096--72109.

\bibitem[{Hurst et~al.(2024)Hurst, Lerer, Goucher, Perelman, Ramesh, Clark,
  Ostrow, Welihinda, Hayes, Radford et~al.}]{gpt-4o}
Aaron Hurst, Adam Lerer, Adam~P Goucher, Adam Perelman, Aditya Ramesh, Aidan
  Clark, AJ~Ostrow, Akila Welihinda, Alan Hayes, Alec Radford, and 1 others.
  2024.
\newblock Gpt-4o system card.
\newblock \emph{arXiv preprint arXiv:2410.21276}.

\bibitem[{Kwon et~al.(2023)Kwon, Li, Zhuang, Sheng, Zheng, Yu, Gonzalez, Zhang,
  and Stoica}]{vllm}
Woosuk Kwon, Zhuohan Li, Siyuan Zhuang, Ying Sheng, Lianmin Zheng, Cody~Hao Yu,
  Joseph~E. Gonzalez, Hao Zhang, and Ion Stoica. 2023.
\newblock Efficient memory management for large language model serving with
  pagedattention.
\newblock In \emph{Proceedings of the ACM SIGOPS 29th Symposium on Operating
  Systems Principles}.

\bibitem[{Lau et~al.(2018)Lau, Gayen, Ben~Abacha, and Demner-Fushman}]{vqa-rad}
Jason~J Lau, Soumya Gayen, Asma Ben~Abacha, and Dina Demner-Fushman. 2018.
\newblock A dataset of clinically generated visual questions and answers about
  radiology images.
\newblock \emph{Scientific data}, 5(1):1--10.

\bibitem[{Lee et~al.(2024)Lee, Wattanawong, Kim, Mangalam, Shen, Anumanchipali,
  Mahoney, Keutzer, and Gholami}]{llm2llm}
Nicholas Lee, Thanakul Wattanawong, Sehoon Kim, Karttikeya Mangalam, Sheng
  Shen, Gopala Anumanchipali, Michael~W Mahoney, Kurt Keutzer, and Amir
  Gholami. 2024.
\newblock Llm2llm: Boosting llms with novel iterative data enhancement.
\newblock \emph{arXiv preprint arXiv:2403.15042}.

\bibitem[{Li et~al.(2023{\natexlab{a}})Li, Ge, Li, and Shan}]{vilt-survey}
Chen Li, Yixiao Ge, Dian Li, and Ying Shan. 2023{\natexlab{a}}.
\newblock Vision-language instruction tuning: A review and analysis.
\newblock \emph{arXiv preprint arXiv:2311.08172}.

\bibitem[{Li et~al.(2024{\natexlab{a}})Li, Wong, Zhang, Usuyama, Liu, Yang,
  Naumann, Poon, and Gao}]{llava-med}
Chunyuan Li, Cliff Wong, Sheng Zhang, Naoto Usuyama, Haotian Liu, Jianwei Yang,
  Tristan Naumann, Hoifung Poon, and Jianfeng Gao. 2024{\natexlab{a}}.
\newblock Llava-med: Training a large language-and-vision assistant for
  biomedicine in one day.
\newblock \emph{Advances in Neural Information Processing Systems}, 36.

\bibitem[{Li et~al.(2020)Li, Jiang, Gu, Peng, Li, Hong, and Tao}]{CLRS}
Haifeng Li, Hao Jiang, Xin Gu, Jian Peng, Wenbo Li, Liang Hong, and Chao Tao.
  2020.
\newblock {CLRS:} continual learning benchmark for remote sensing image scene
  classification.
\newblock \emph{Sensors}, 20(4):1226.

\bibitem[{Li et~al.(2024{\natexlab{b}})Li, Dong, Tang, Wang, Zhang, Huang,
  Huang, Huang, Huang, Zhang et~al.}]{GLAN}
Haoran Li, Qingxiu Dong, Zhengyang Tang, Chaojun Wang, Xingxing Zhang, Haoyang
  Huang, Shaohan Huang, Xiaolong Huang, Zeqiang Huang, Dongdong Zhang, and 1
  others. 2024{\natexlab{b}}.
\newblock Synthetic data (almost) from scratch: Generalized instruction tuning
  for language models.
\newblock \emph{arXiv preprint arXiv:2402.13064}.

\bibitem[{Li et~al.(2023{\natexlab{b}})Li, Yu, Zhou, Schick, Levy, Zettlemoyer,
  Weston, and Lewis}]{selfalignment}
Xian Li, Ping Yu, Chunting Zhou, Timo Schick, Omer Levy, Luke Zettlemoyer,
  Jason~E Weston, and Mike Lewis. 2023{\natexlab{b}}.
\newblock Self-alignment with instruction backtranslation.
\newblock In \emph{The Twelfth International Conference on Learning
  Representations}.

\bibitem[{Liu et~al.(2021)Liu, Zhan, Xu, Ma, Yang, and Wu}]{slake}
Bo~Liu, Li-Ming Zhan, Li~Xu, Lin Ma, Yan Yang, and Xiao-Ming Wu. 2021.
\newblock Slake: A semantically-labeled knowledge-enhanced dataset for medical
  visual question answering.
\newblock In \emph{2021 IEEE 18th International Symposium on Biomedical Imaging
  (ISBI)}, pages 1650--1654. IEEE.

\bibitem[{Liu et~al.(2024{\natexlab{a}})Liu, Zhang, Qiu, Huang, Lin, Zhao,
  Geng, Lin, Jin, Zhang, Shao, Xu, He, He, Shao, Lu, Qiao, Li, and
  Gao}]{SPHINX-X}
Dongyang Liu, Renrui Zhang, Longtian Qiu, Siyuan Huang, Weifeng Lin, Shitian
  Zhao, Shijie Geng, Ziyi Lin, Peng Jin, Kaipeng Zhang, Wenqi Shao, Chao Xu,
  Conghui He, Junjun He, Hao Shao, Pan Lu, Yu~Qiao, Hongsheng Li, and Peng Gao.
  2024{\natexlab{a}}.
\newblock {SPHINX-X:} scaling data and parameters for a family of multi-modal
  large language models.
\newblock In \emph{{ICML}}. OpenReview.net.

\bibitem[{Liu et~al.(2024{\natexlab{b}})Liu, Li, Li, Li, Zhang, Shen, and
  Lee}]{llava-v1.6}
Haotian Liu, Chunyuan Li, Yuheng Li, Bo~Li, Yuanhan Zhang, Sheng Shen, and
  Yong~Jae Lee. 2024{\natexlab{b}}.
\newblock Llava-next: Improved reasoning, ocr, and world knowledge.

\bibitem[{Liu et~al.(2024{\natexlab{c}})Liu, Li, Wu, and Lee}]{llava}
Haotian Liu, Chunyuan Li, Qingyang Wu, and Yong~Jae Lee. 2024{\natexlab{c}}.
\newblock Visual instruction tuning.
\newblock \emph{Advances in neural information processing systems}, 36.

\bibitem[{Liu et~al.(2024{\natexlab{d}})Liu, Wei, Liu, Si, Zhang, Rao, Zheng,
  Peng, Yang, Zhou et~al.}]{syntheticsurvey}
Ruibo Liu, Jerry Wei, Fangyu Liu, Chenglei Si, Yanzhe Zhang, Jinmeng Rao,
  Steven Zheng, Daiyi Peng, Diyi Yang, Denny Zhou, and 1 others.
  2024{\natexlab{d}}.
\newblock Best practices and lessons learned on synthetic data for language
  models.
\newblock \emph{arXiv preprint arXiv:2404.07503}.

\bibitem[{Lu et~al.()Lu, Wang, Zheng, and Li}]{rsicd}
Xiaoqiang Lu, Binqiang Wang, Xiangtao Zheng, and Xuelong Li.
\newblock \href {https://doi.org/10.1109/TGRS.2017.2776321} {Exploring models
  and data for remote sensing image caption generation}.
\newblock \emph{IEEE Transactions on Geoscience and Remote Sensing},
  56(4):2183--2195.

\bibitem[{Luo et~al.(2023{\natexlab{a}})Luo, Zhang, Fan, Yang, Wu, Qiao, and
  Nie}]{biomedgpt}
Yizhen Luo, Jiahuan Zhang, Siqi Fan, Kai Yang, Yushuai Wu, Mu~Qiao, and Zaiqing
  Nie. 2023{\natexlab{a}}.
\newblock Biomedgpt: Open multimodal generative pre-trained transformer for
  biomedicine.
\newblock \emph{arXiv preprint arXiv:2308.09442}.

\bibitem[{Luo et~al.(2023{\natexlab{b}})Luo, Yang, Meng, Li, Zhou, and
  Zhang}]{catastrophic-forget}
Yun Luo, Zhen Yang, Fandong Meng, Yafu Li, Jie Zhou, and Yue Zhang.
  2023{\natexlab{b}}.
\newblock An empirical study of catastrophic forgetting in large language
  models during continual fine-tuning.
\newblock \emph{arXiv preprint arXiv:2308.08747}.

\bibitem[{Ma et~al.(2024)Ma, Cao, Sun, Pavone, and Xiao}]{dolphins}
Yingzi Ma, Yulong Cao, Jiachen Sun, Marco Pavone, and Chaowei Xiao. 2024.
\newblock Dolphins: Multimodal language model for driving.
\newblock In \emph{{ECCV} {(45)}}, volume 15103 of \emph{Lecture Notes in
  Computer Science}, pages 403--420. Springer.

\bibitem[{Mohbat and Zaki(2024)}]{llava-chef}
Fnu Mohbat and Mohammed~J Zaki. 2024.
\newblock Llava-chef: A multi-modal generative model for food recipes.
\newblock In \emph{Proceedings of the 33rd ACM International Conference on
  Information and Knowledge Management}, pages 1711--1721.

\bibitem[{Moor et~al.(2023)Moor, Huang, Wu, Yasunaga, Dalmia, Leskovec, Zakka,
  Reis, and Rajpurkar}]{medflamigo}
Michael Moor, Qian Huang, Shirley Wu, Michihiro Yasunaga, Yash Dalmia, Jure
  Leskovec, Cyril Zakka, Eduardo~Pontes Reis, and Pranav Rajpurkar. 2023.
\newblock Med-flamingo: a multimodal medical few-shot learner.
\newblock In \emph{Machine Learning for Health (ML4H)}, pages 353--367. PMLR.

\bibitem[{Mukherjee et~al.(2023)Mukherjee, Mitra, Jawahar, Agarwal, Palangi,
  and Awadallah}]{orca}
Subhabrata Mukherjee, Arindam Mitra, Ganesh Jawahar, Sahaj Agarwal, Hamid
  Palangi, and Ahmed Awadallah. 2023.
\newblock Orca: Progressive learning from complex explanation traces of gpt-4.
\newblock \emph{arXiv preprint arXiv:2306.02707}.

\bibitem[{OpenAI(2023)}]{gpt-4v}
OpenAI. 2023.
\newblock \href {https://api.semanticscholar.org/CorpusID:263218031}
  {Gpt-4v(ision) system card}.

\bibitem[{Peng et~al.(2024)Peng, Wang, Dong, Hao, Huang, Ma, Ye, and
  Wei}]{kosmos2}
Zhiliang Peng, Wenhui Wang, Li~Dong, Yaru Hao, Shaohan Huang, Shuming Ma,
  Qixiang Ye, and Furu Wei. 2024.
\newblock Grounding multimodal large language models to the world.
\newblock In \emph{The Twelfth International Conference on Learning
  Representations}.

\bibitem[{Qu et~al.(2016)Qu, Li, Tao, and Lu}]{syndney}
Bo~Qu, Xuelong Li, Dacheng Tao, and Xiaoqiang Lu. 2016.
\newblock Deep semantic understanding of high resolution remote sensing image.
\newblock In \emph{{CITS}}, pages 1--5. {IEEE}.

\bibitem[{Radford et~al.(2021)Radford, Kim, Hallacy, Ramesh, Goh, Agarwal,
  Sastry, Askell, Mishkin, Clark et~al.}]{clip}
Alec Radford, Jong~Wook Kim, Chris Hallacy, Aditya Ramesh, Gabriel Goh,
  Sandhini Agarwal, Girish Sastry, Amanda Askell, Pamela Mishkin, Jack Clark,
  and 1 others. 2021.
\newblock Learning transferable visual models from natural language
  supervision.
\newblock In \emph{International conference on machine learning}, pages
  8748--8763. PMLR.

\bibitem[{Radford et~al.(2018)Radford, Narasimhan, Salimans, Sutskever
  et~al.}]{gpt1}
Alec Radford, Karthik Narasimhan, Tim Salimans, Ilya Sutskever, and 1 others.
  2018.
\newblock Improving language understanding by generative pre-training.

\bibitem[{Rahnemoonfar et~al.(2021)Rahnemoonfar, Chowdhury, Sarkar, Varshney,
  Yari, and Murphy}]{FloodNet}
Maryam Rahnemoonfar, Tashnim Chowdhury, Argho Sarkar, Debvrat Varshney, Masoud
  Yari, and Robin~Roberson Murphy. 2021.
\newblock Floodnet: {A} high resolution aerial imagery dataset for post flood
  scene understanding.
\newblock \emph{{IEEE} Access}, 9:89644--89654.

\bibitem[{Salvador et~al.(2017)Salvador, Hynes, Aytar, Marin, Ofli, Weber, and
  Torralba}]{recipe1m}
Amaia Salvador, Nicholas Hynes, Yusuf Aytar, Javier Marin, Ferda Ofli, Ingmar
  Weber, and Antonio Torralba. 2017.
\newblock Learning cross-modal embeddings for cooking recipes and food images.
\newblock In \emph{Proceedings of the IEEE conference on computer vision and
  pattern recognition}, pages 3020--3028.

\bibitem[{Team(2024)}]{Chameleon}
Chameleon Team. 2024.
\newblock Chameleon: Mixed-modal early-fusion foundation models.
\newblock \emph{CoRR}, abs/2405.09818.

\bibitem[{Thames et~al.(2021)Thames, Karpur, Norris, Xia, Panait, Weyand, and
  Sim}]{nutrition5k}
Quin Thames, Arjun Karpur, Wade Norris, Fangting Xia, Liviu Panait, Tobias
  Weyand, and Jack Sim. 2021.
\newblock Nutrition5k: Towards automatic nutritional understanding of generic
  food.
\newblock In \emph{Proceedings of the IEEE/CVF Conference on Computer Vision
  and Pattern Recognition}, pages 8903--8911.

\bibitem[{Touvron et~al.(2023)Touvron, Lavril, Izacard, Martinet, Lachaux,
  Lacroix, Rozi{\`e}re, Goyal, Hambro, Azhar et~al.}]{llama}
Hugo Touvron, Thibaut Lavril, Gautier Izacard, Xavier Martinet, Marie-Anne
  Lachaux, Timoth{\'e}e Lacroix, Baptiste Rozi{\`e}re, Naman Goyal, Eric
  Hambro, Faisal Azhar, and 1 others. 2023.
\newblock Llama: Open and efficient foundation language models.
\newblock \emph{arXiv preprint arXiv:2302.13971}.

\bibitem[{Wang et~al.(2024)Wang, Bai, Tan, Wang, Fan, Bai, Chen, Liu, Wang, Ge
  et~al.}]{qwen2-vl}
Peng Wang, Shuai Bai, Sinan Tan, Shijie Wang, Zhihao Fan, Jinze Bai, Keqin
  Chen, Xuejing Liu, Jialin Wang, Wenbin Ge, and 1 others. 2024.
\newblock Qwen2-vl: Enhancing vision-language model's perception of the world
  at any resolution.
\newblock \emph{arXiv preprint arXiv:2409.12191}.

\bibitem[{Wang et~al.(2022)Wang, Wei, Schuurmans, Le, Chi, Narang, Chowdhery,
  and Zhou}]{Self-consistency}
Xuezhi Wang, Jason Wei, Dale Schuurmans, Quoc Le, Ed~Chi, Sharan Narang,
  Aakanksha Chowdhery, and Denny Zhou. 2022.
\newblock Self-consistency improves chain of thought reasoning in language
  models.
\newblock \emph{arXiv preprint arXiv:2203.11171}.

\bibitem[{Wei et~al.(2024)Wei, Zhong, Tan, Zeng, Liu, Zhao, and
  Yang}]{InstructSeg}
Cong Wei, Yujie Zhong, Haoxian Tan, Yingsen Zeng, Yong Liu, Zheng Zhao, and
  Yujiu Yang. 2024.
\newblock Instructseg: Unifying instructed visual segmentation with multi-modal
  large language models.
\newblock \emph{CoRR}, abs/2412.14006.

\bibitem[{Wei et~al.(2021)Wei, Bosma, Zhao, Guu, Yu, Lester, Du, Dai, and
  Le}]{flan}
Jason Wei, Maarten Bosma, Vincent~Y Zhao, Kelvin Guu, Adams~Wei Yu, Brian
  Lester, Nan Du, Andrew~M Dai, and Quoc~V Le. 2021.
\newblock Finetuned language models are zero-shot learners.
\newblock \emph{arXiv preprint arXiv:2109.01652}.

\bibitem[{Wei et~al.(2022)Wei, Wang, Schuurmans, Bosma, Xia, Chi, Le, Zhou
  et~al.}]{cot}
Jason Wei, Xuezhi Wang, Dale Schuurmans, Maarten Bosma, Fei Xia, Ed~Chi, Quoc~V
  Le, Denny Zhou, and 1 others. 2022.
\newblock Chain-of-thought prompting elicits reasoning in large language
  models.
\newblock \emph{Advances in neural information processing systems},
  35:24824--24837.

\bibitem[{Wu et~al.(2021)Wu, Fu, Liu, Lim, Hoi, and Sun}]{foodseg103}
Xiongwei Wu, Xin Fu, Ying Liu, Ee-Peng Lim, Steven~CH Hoi, and Qianru Sun.
  2021.
\newblock A large-scale benchmark for food image segmentation.
\newblock In \emph{Proceedings of the 29th ACM international conference on
  multimedia}, pages 506--515.

\bibitem[{Xu et~al.(2023)Xu, Sun, Zheng, Geng, Zhao, Feng, Tao, Lin, and
  Jiang}]{wizardlm}
Can Xu, Qingfeng Sun, Kai Zheng, Xiubo Geng, Pu~Zhao, Jiazhan Feng, Chongyang
  Tao, Qingwei Lin, and Daxin Jiang. 2023.
\newblock Wizardlm: Empowering large pre-trained language models to follow
  complex instructions.
\newblock In \emph{The Twelfth International Conference on Learning
  Representations}.

\bibitem[{Xu et~al.(2024)Xu, Feng, Shao, Ashby, Shen, Jin, Cheng, Wang, and
  Huang}]{vision-flan}
Zhiyang Xu, Chao Feng, Rulin Shao, Trevor Ashby, Ying Shen, Di~Jin, Yu~Cheng,
  Qifan Wang, and Lifu Huang. 2024.
\newblock Vision-flan: Scaling human-labeled tasks in visual instruction
  tuning.
\newblock \emph{arXiv preprint arXiv:2402.11690}.

\bibitem[{Yang and Newsam(2010)}]{UCMerced}
Yi~Yang and Shawn Newsam. 2010.
\newblock Bag-of-visual-words and spatial extensions for land-use
  classification.
\newblock In \emph{ACM SIGSPATIAL International Conference on Advances in
  Geographic Information Systems (ACM GIS)}.

\bibitem[{Yin et~al.(2023)Yin, Qi, Zhu, Chen, Jiang, and Ngo}]{foodlmm}
Yuehao Yin, Huiyan Qi, Bin Zhu, Jingjing Chen, Yu-Gang Jiang, and Chong-Wah
  Ngo. 2023.
\newblock Foodlmm: A versatile food assistant using large multi-modal model.
\newblock \emph{arXiv preprint arXiv:2312.14991}.

\bibitem[{Yuan et~al.(2022)Yuan, Zhang, Fu, Li, Deng, Wang, and Sun}]{rsitmd}
Zhiqiang Yuan, Wenkai Zhang, Kun Fu, Xuan Li, Chubo Deng, Hongqi Wang, and Xian
  Sun. 2022.
\newblock Exploring a fine-grained multiscale method for cross-modal remote
  sensing image retrieval.
\newblock \emph{{IEEE} Trans. Geosci. Remote. Sens.}, 60:1--19.

\bibitem[{Yue et~al.(2024)Yue, Zheng, Zhang, and Chen}]{mammoth2}
Xiang Yue, Tuney Zheng, Ge~Zhang, and Wenhu Chen. 2024.
\newblock Mammoth2: Scaling instructions from the web.
\newblock \emph{arXiv preprint arXiv:2405.03548}.

\bibitem[{Zhang et~al.(2023{\natexlab{a}})Zhang, Xu, Usuyama, Bagga, Tinn,
  Preston, Rao, Wei, Valluri, Wong et~al.}]{pmc-15m}
Sheng Zhang, Yanbo Xu, Naoto Usuyama, Jaspreet Bagga, Robert Tinn, Sam Preston,
  Rajesh Rao, Mu~Wei, Naveen Valluri, Cliff Wong, and 1 others.
  2023{\natexlab{a}}.
\newblock Large-scale domain-specific pretraining for biomedical
  vision-language processing.
\newblock \emph{arXiv preprint arXiv:2303.00915}, 2(3):6.

\bibitem[{Zhang et~al.(2024)Zhang, Cai, Zhang, Zhuang, and Mao}]{earthgpt}
Wei Zhang, Miaoxin Cai, Tong Zhang, Yin Zhuang, and Xuerui Mao. 2024.
\newblock Earthgpt: A universal multi-modal large language model for
  multi-sensor image comprehension in remote sensing domain.
\newblock \emph{IEEE Transactions on Geoscience and Remote Sensing}.

\bibitem[{Zhang et~al.(2023{\natexlab{b}})Zhang, Wu, Zhao, Lin, Zhang, Wang,
  and Xie}]{pmc-vqa}
Xiaoman Zhang, Chaoyi Wu, Ziheng Zhao, Weixiong Lin, Ya~Zhang, Yanfeng Wang,
  and Weidi Xie. 2023{\natexlab{b}}.
\newblock Pmc-vqa: Visual instruction tuning for medical visual question
  answering.
\newblock \emph{arXiv preprint arXiv:2305.10415}.

\bibitem[{Zhao et~al.(2023)Zhao, Zhou, and Shou}]{genixer}
Henry~Hengyuan Zhao, Pan Zhou, and Mike~Zheng Shou. 2023.
\newblock Genixer: Empowering multimodal large language models as a powerful
  data generator.
\newblock \emph{arXiv preprint arXiv:2312.06731}.

\bibitem[{Zheng et~al.(2024)Zheng, Zhang, Zhang, Ye, Luo, Feng, and
  Ma}]{llamafactory}
Yaowei Zheng, Richong Zhang, Junhao Zhang, Yanhan Ye, Zheyan Luo, Zhangchi
  Feng, and Yongqiang Ma. 2024.
\newblock \href {http://arxiv.org/abs/2403.13372} {Llamafactory: Unified
  efficient fine-tuning of 100+ language models}.
\newblock In \emph{Proceedings of the 62nd Annual Meeting of the Association
  for Computational Linguistics (Volume 3: System Demonstrations)}, Bangkok,
  Thailand. Association for Computational Linguistics.

\end{thebibliography}

\clearpage

\appendix

\section{Extended Related Work}
\label{app:Extended Related Work}
\paragraph{Text-only/Visual-language Instruction Synthesis}

For text-only instruction synthesis, the most notable distinction between our work and these previous works is our inclusion of images as an additional source of information. In addition to the input modality, our method also differs in several key aspects. Instead of distilling knowledge from strong models~\citep{wizardlm,orca,GLAN}, we focus on learning from image-caption sources. Additionally, we outperform rule-based methods~\citep{adaptllm,udit} by increasing instruction diversity. Iterative techniques~\citep{selfalignment,mammoth2,llm2llm} could complement our method. Among all text-only instruction synthesis works, we draw the most inspiration from Instruct Pre-Training~\citep{instructpt}. However, we introduce a consistency filter that significantly improves accuracy and reduces the need for domain expert annotation, allowing us to surpass synthesizers based on closed-source models. Additionally, we incorporate several design choices to balance the generalization challenge between image and text inputs.

For visual-language instruction synthesis, using closed-source/open-source models to generate data is common in general MLLM training~\citep{vilt-survey}. Notable works include LLaVA~\citep{llava}, which uses text-only GPT-4 to generate instructions based on captions as if the model could ``see'' the image, leading to exceptional general-task performance. Additionally, ALLaVA~\citep{allava} leverages GPT-4V to generate large-scale visual instructions from image-caption pairs while also augmenting answers from Vision-FLAN~\citep{vision-flan}. However, for domain-specific MLLMs, particularly in private domains, closed-source models pose privacy concerns. Thus, we focus on open-source models for instruction synthesis. Even with this focus, our method still outperforms LLaVA-Med (which uses text-only GPT-4) and PubMedVision (which uses GPT-4V), as shown in Section~\ref{sec:Main Results}, due to our superior utilization of domain knowledge, task diversity, and complexity (discussed in Section~\ref{sec:Analysis}). The most relevant open-source model approach is Genxier~\citep{genixer}, which fine-tunes open-source MLLMs to synthesize instructions from images. However, whereas Genxier focuses on general training, our work is specifically dedicated to domain-specific adaptation. Moreover, we reproduce a Genxier-like baseline (Column 2, Table~\ref{tab:synthesizer}), where image-only utilization underperforms our method.

\paragraph{Multi/Single-Stage MLLM Training} The training of general MLLMs typically starts with an unaligned LLM~\citep{gpt3,palm,llama} and visual encoder~\citep{clip}, and often proceeds in two stages. One representative example is LLaVA~\citep{llava}, which first trains on image-caption pairs, and then on visual instructions. Note that ``native multimodal language models" also exist, such as Kosmos~\citep{kosmos1,kosmos2} and Chameleon~\citep{Chameleon}, where all modalities are trained end-to-end from scratch. In this paper, we mainly focus on the first type which is more commonly used, possibly due to its training efficiency which avoids the need for pre-training LLMs. In addition to multi-stage training, some works have explored the benefits of single-stage training, such as SPHINX-X~\citep{SPHINX-X} and InstructSeg~\citep{InstructSeg}. However, the primary motivation behind SPHINX-X is to simplify the process by eliminating the intensive effort of assigning tunable parameters and dataset combinations to different stages. Furthermore, neither SPHINX-X nor InstructSeg provides the detailed comparison of the impacts on downstream tasks with different training strategies that we present. Additionally, two-stage training remains the mainstream approach for domain-specific training of MLLMs.

\section{Seed Data Construction and Distribution}
\label{app:Seed Data}
\begin{figure}[!htb]
    \centering
    \includegraphics[width=\columnwidth]{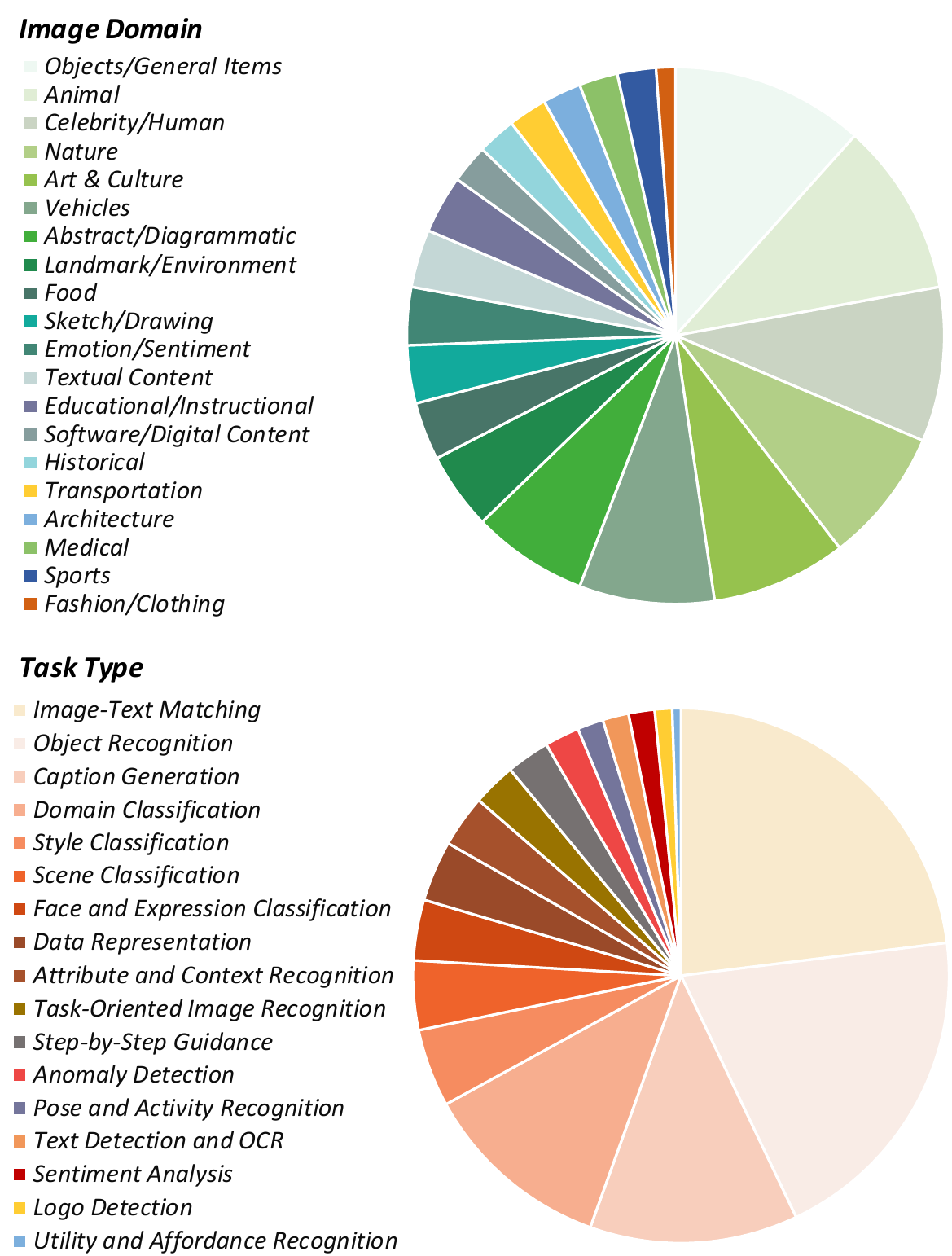}
    \caption{\textbf{Distribution of Image Domains and Task Types in Seed Data.}}
\label{fig:seed_distribution}
\end{figure}

We convert the combination of VisionFLAN~\citep{vision-flan} and ALLaVA~\citep{allava} into our required format. Each seed data example consists of an image-caption pair as the input and a related task triplet as the output, which includes an instruction, an informative response, and a precise response. VisionFLAN is a visual instruction task dataset containing 191 tasks, each with 1K examples. ALLaVA builds on VisionFLAN by generating a caption for each image and regenerating a response for each instruction. In our format, the image, instruction, and response from VisionFLAN are used as the image, instruction, and precise response, respectively, while the caption and re-generated response from ALLaVA are used as the caption and informative response. Benefiting from the diversity of existing datasets, our seed data encompass a wide range of image domains and task types, as shown in Figure~\ref{fig:seed_distribution}.

\section{Image-Caption Data Source}
\label{app:Image-Caption Data Source}
For the biomedicine domain, we use two datasets from PubMed Central under the \texttt{MIT License}: (1) \pmcraw~\citep{pmc-15m}, which comprises 470K publicly available images with human-annotated captions, and (2) \pmcrefine, which contains 510K image-caption pairs with captions refined by an MLLM. For the food domain, we collect 130K single-image examples from \recipe~\citep{recipe1m}, which is licensed under \texttt{CC-BY-4.0}. For the remote sensing domain, we collect 40K image-caption pairs from NWPU-Captions~\citep{NWPU}, RSICD~\citep{rsicd}, RSITMD~\citep{rsitmd}, Sydney-Captions~\citep{syndney}, and UCM-Captions~\citep{syndney}.

\section{Implementations and Costs of Visual Instruction Synthesis}
\label{app:Visual Instruction Synthesizer}
Table~\ref{tab:Hyper-Parameters for Synthesizer Tuning} presents the hyper-parameters used for synthesizer tuning. We employ the vLLM inference framework~\cite{vllm} to speed up task synthesis and consistency checks. On a single A100-80GB GPU, it takes approximately 10 hours to synthesize task triplets and an additional 2.5 hours to perform consistency-based filtering for every 100K image-caption pairs. On average, about 30\% of the task triplets are reserved after filtering. Specifically, we collect 150K, 144K, and 32K, 15K instruction-response pairs for \pmcraw, \pmcrefine, \recipe~and remote sensing image-caption pairs, respectively.

\begin{table}[!htb]
\centering
\resizebox{0.82\columnwidth}{!}{\begin{tabular}{ll}
\toprule
 \textbf{Hyper-Parameter}              & \textbf{Assignment}                                      \\ \midrule
Base Model     & LLaVA-v1.6-8B   \\ 
Trainable  & Full Model \\
Epoch      & 2      \\
Batch Size        & 128  \\
Max Seq Length      & 6144 \\
LR$_{\text{projector~\&~LLM}}$               & 2e-5   \\
LR$_{\text{visual~encoder}}$               & 2e-6   \\
LR Scheduler                & Cosine  \\
Weight Decay & 0     \\
Warm-Up Ratio              & 0.03  \\
Computing Infrastructure & 8 A100-80GB GPUs          \\
Training Time & 13 Hours \\ \bottomrule
\end{tabular}%
}
\caption{\textbf{Hyper-Parameters for Synthesizer Tuning}}
\label{tab:Hyper-Parameters for Synthesizer Tuning}
\end{table}

\begin{table}[!htb]
\centering
\resizebox{\columnwidth}{!}{\begin{tabular}{llll}
\toprule

\textbf{MLLM} & \textbf{LLaVA-v1.6}                      & \textbf{Qwen2-VL}                   & \textbf{Llama-3.2}                       \\ \midrule
Trainable  & Full Model                  & Full Model             & Full Model        \\
Epoch      & 1                     & 1                 & 1      \\
Batch Size        & 128                     & 128     & 128                   \\
Max Seq Length      & 6144                      & 6144 & 6144                        \\
LR$_{\text{projector~\&~LLM}}$               & 2e-5     & 1e-5              & 5e-6                   \\
LR$_{\text{visual~encoder}}$              & 2e-6      & 1e-5            & 5e-6                   \\
LR Scheduler                & Cosine                & Cosine                   & Cosine                   \\
Weight Decay & 0       & 0.1  & 0.1         \\
Warm-Up Ratio              & 0.03                   & 0.1  & 0.1                       \\\bottomrule
\end{tabular}%
}
\caption{\textbf{Hyper-Parameters for MLLM Single-Stage Post-Training.}}
\label{tab:Hyper-Parameters for MLLM Post-Training}
\end{table}

\begin{table}[!htb]
\centering
\resizebox{\columnwidth}{!}{\begin{tabular}{lllll}
\toprule
 \textbf{Image-Caption} & \textbf{\pmcraw} & \textbf{\pmcrefine} & \textbf{\recipe} & \textbf{Remote.}\\ \midrule
LLaVA-v1.6-8B         & 21                     & 23                         & 6          & 2              \\
Qwen2-VL-2B           & 3.5                     & 4                         & 1         & 0.5            \\
Llama-3.2-11B         & 29                     & 31                         & 9     &  3 \\ \bottomrule               
\end{tabular}
}
\caption{\textbf{Training Time (Hours) for MLLM Single-Stage Post-Training} on 8 A100-80GB GPUs.}
\label{tab:training cost}
\end{table}

\section{MLLM Post-Training Settings and Costs}
\label{app:MLLM Post-Training Settings}
Tables~\ref{tab:Hyper-Parameters for MLLM Post-Training} and~\ref{tab:training cost} present the hyper-parameters and training time for the single-stage post-training of MLLMs. In the two-stage training experiments, we employ two common approaches for the first stage on image-caption pairs: (1) unfreezing only the vision-language projector~\citep{llava} for {LLaVA-v1.6-8B}, and (2) unfreezing the full model~\citep{sharegpt4v} for {Qwen2-VL-2B} and {Llama-3.2-11B}. During the second stage on visual instruction tasks, we unfreeze the full model across all setups. The hyperparameters for both stages on Qwen2-VL-2B and Llama-3.2-11B are the same as those listed in the table. For LLaVA-v1.6-8B, the first stage differs only in the trainable module, which is the vision-language projector, using a learning rate of 2e-3. The hyper-parameters for the second stage are the same as those listed in Table~\ref{tab:Hyper-Parameters for MLLM Post-Training}. We use the code implementations from~\citep{open-llava-next} for experiments on LLaVA-v1.6-8B and from~\citep{llamafactory} for experiments on Qwen2-VL-2B and Llama-3.2-11B.

\section{Further Analysis on Fine-tuning with Blank Images}
\label{app:Further Ablations in Replacing Images with a Blank One}
In our initial synthesizer fine-tuning using 100\% intact image-caption pairs, we observe two key limitations when testing on out-of-distribution images: (1) the synthesizer generates overly simple questions with low task diversity, and (2) it largely ignores caption information despite its rich knowledge content. Upon examining the seed data source—VisionFlan~\citep{vision-flan}, we find that the task instructions are annotated based solely on images without using captions. This explains why our synthesizer learns to prioritize visual information while neglecting textual input.

To balance the utilization of both modalities, we introduce blank images by replacing 10\% of training images. This modification forces the synthesizer to rely on caption information when visual information becomes unavailable. Empirical results (Table~\ref{tab:ablation} in Section~\ref{sec:ablations} and Table~\ref{tab:synthesizer} in Section~\ref{sec:Domain-Specific Visual Instruction Synthesis}) demonstrate the effectiveness of this strategy. We further validate our design through robustness testing with corrupted inputs. Table~\ref{tab:blank_image_ablation} reveals that fine-tuning with blank images produces two key benefits: first, it yields significant quality improvements for corrupted images while maintaining comparable performance on intact inputs; second, it preserves task quality for text-corrupted cases. These results demonstrate that our approach successfully balances multimodal utilization while enhancing robustness to challenging visual inputs.

\begin{table}[!htb]
\centering
\resizebox{\columnwidth}{!}{
\begin{tabular}{lccc}
\toprule
\diagbox[width=2.9cm]{\textbf{Fine-Tune}}{\textbf{Test}} & \textbf{Intact} & \textbf{Corrupt Image} & \textbf{Corrupt Caption} \\
\midrule
w/o Blank & 45.0 & 32.2 & 38.9 \\
w/ Blank & 44.7 & 36.1 & 38.9 \\
\bottomrule
\end{tabular}
}
\caption{\textbf{Synthetic Task Quality Comparison} between synthesizers fine-tuned with/without blank images (10\% replacement).``Intact'' uses original image-caption pairs; ``Corrupt Image'' applies random noise or crops to images; ``Corrupt Caption'' randomly removes text segments from captions.}
\label{tab:blank_image_ablation}
\end{table}

\section{Scoring Criteria for Data Quality}
\label{app:Scoring Criteria for Data Quality}

For each synthetic dataset, we sample 200 examples and use the following scoring criteria to evaluate data quality in each aspect. The final scores are rescaled to a 0-100 range for presentation uniformity.

\paragraph{Task Diversity} For each instruction-response pair, the annotator selects the most appropriate category from the common vision instruction task types listed below. Once all data samples are annotated, we report the number of distinct task types normalized by the total number of common task types.

\begin{itemize}[leftmargin=*]
\itemsep0em
\item \textit{Domain Classification}: Classifying images into domains like race, animal categories, and environment types.
\item \textit{Object Recognition}: Recognizing detailed objects like animal species, car brands, and specific object types.
\item \textit{Pose and Activity Recognition}: Identifying specific human poses and activities.
\item \textit{Logo Detection}: Detecting and recognizing brand logos.
\item \textit{Face and Expression Classification}: Classifying facial attributes by age, gender, and detecting expressions.
\item \textit{Scene Classification}: Categorizing images into scene types like beaches, forests, and cities.
\item \textit{Sentiment Analysis}: Detecting sentiment in images.
\item \textit{Caption Generation}: Generating captions for images, including general and contextual descriptions.
\item \textit{Text Detection and OCR}: Recognizing text in images and structured text detection.
\item \textit{Image-Text Matching}: Assessing image-text similarity and coherence for multimodal content.
\item \textit{Anomaly Detection}: Identifying anomalies in settings like industrial and road scenes.
\item \textit{Style Classification}: Classifying images by artistic style and quality.
\item \textit{Attribute and Context Recognition}: Detecting image attributes and contexts, such as object presence and temporal classification.
\item \textit{Task-Oriented Image Recognition}: Recognizing objects in structured contexts, like weed species and quick-draw sketches.
\item \textit{Step-by-Step Guidance}: Recognizing steps in instructional content, like wikihow procedures.
\item \textit{Data Representation and Visualization}: Visual QA for charts and chart captioning.
\item \textit{Utility and Affordance Recognition}: Detecting object utility or affordance in images.
\item \textit{Visual Grounding}: Linking image parts to corresponding words or phrases.
\item \textit{Segmentation}: Dividing images into meaningful segments, identifying objects or regions.
\item \textit{Visual Storytelling}: Creating narratives based on a series of images.
\end{itemize}

\paragraph{Domain Knowledge Utilization} For each instruction-response pair, the annotator evaluates the extent to which domain-specific knowledge from the image is utilized to complete the task. The scoring follows the criteria below, and we report the average score across all samples.

\begin{itemize}[leftmargin=*]
\itemsep0em
\item 1: The task is totally irrelevant to the image.
\item 2: The task is relevant, but the question is mundane and answerable without reviewing the image.
\item 3: The task requires reviewing the image, but the question is vague, such as asking for a general caption.
\item 4: The task is clear, but the question focuses on only one detail in the image.
\item 5: The task is highly relevant to both the details and overall context of the image.
\end{itemize}

\paragraph{Task Complexity} For each instruction-response pair, the annotator assesses task complexity, with higher scores for tasks requiring reasoning and instruction-following abilities, using the criteria below. We report the average score across all samples.

\begin{itemize}[leftmargin=*]
\itemsep0em
\item 1: The task can be easily completed by mimicking part of the caption.
\item 2: The task can be easily completed by reviewing the image, such as identifying an obvious object.
\item 3: The task requires consideration of the details.
\item 4: The task requires complex reasoning on details and overview.
\item 5: The task requires complex reasoning and instruction-following abilities, such as returning the answer in a required format.
\end{itemize}

\paragraph{Response Accuracy} For each instruction-response pair, the annotator assesses whether the response correctly addresses the task based on the context, using the following criteria. We report the average score across all samples.

\begin{itemize}[leftmargin=*]
\itemsep0em
\item 1: The response is totally irrelevant to the task instruction.
\item 2: The response attempts to address the instruction, but both the reasoning and conclusion are incorrect.
\item 3: The reasoning is correct, but the conclusion is incorrect.
\item 4: The conclusion is correct, but the reasoning is incorrect.
\item 5: Both the reasoning and conclusion are correct.
\end{itemize}

\section{Task Evaluation Details}
\label{app:Task Evaluation Details}
Tables~\ref{tab:tasks} and~\ref{tab:tasks_template} present the specifications and prompt templates for evaluated tasks in each domain. We conduct zero-shot prompting evaluations on these tasks.

\textbf{For biomedicine}, we follow the evaluation approach of~\citep{llava-med} for SLAKE, PathVQA, and VQA-RAD, and the method of~\citep{pmc-vqa} for PMC-VQA.
\begin{itemize}[leftmargin=*]
\itemsep0em
\item \textit{SLAKE}~\citep{slake} is a semantically-labeled, knowledge-enhanced medical VQA dataset with radiology images and diverse QA pairs annotated by physicians. The dataset includes semantic segmentation masks, object detection bounding boxes, and covers various body parts. ``CLOSED'' answers are yes/no type, while ``OPEN'' answers are one-word or short phrases. We use only the English subset.
\item \textit{PathVQA}~\citep{pathvqa} consists of pathology images with QA pairs covering aspects like location, shape, and color. Questions are categorized as ``OPEN'' (open-ended) or ``CLOSED'' (closed-ended).
\item \textit{VQA-RAD}~\citep{vqa-rad} includes clinician-generated QA pairs and radiology images spanning the head, chest, and abdomen. Questions are categorized into 11 types, with answers as either ``OPEN'' (short text) or ``CLOSED'' (yes/no).
\item \textit{PMC-VQA}~\citep{pmc-vqa} is larger and more diverse MedVQA datasets, with questions ranging from identifying modalities and organs to complex questions requiring specialized knowledge. All questions are multiple-choice.
\end{itemize}

\textbf{For food domain}, the task descriptions are as follows:
\begin{itemize}[leftmargin=*]
\itemsep0em
\item \textit{Recipe1M}~\citep{recipe1m} contains recipe information, including titles, ingredients, and cooking instructions. We evaluate models by taking an image and asking for the recipe name, ingredients, and steps.
\item \textit{Nutrition5K}~\citep{nutrition5k} comprises real-world food dishes with RGB images and nutritional content annotations. We use the ingredient information to create an ingredient prediction task, where the model generates ingredients from an image.
\item \textit{Food101}~\citep{food101} features images across 101 food categories. We ask the model to classify each image into one of the 101 categories.
\item \textit{FoodSeg103}~\citep{foodseg103} includes 103 food categories with images and pixel-wise ingredient annotations. We ask the model to select one or multiple categories from a provided list.
\end{itemize}

\textbf{For remote-sensing domain}, we follow the evaluation approach of~\citep{earthgpt}.
\begin{itemize}[leftmargin=*]
\itemsep0em
\item \textit{CLRS}~\citep{CLRS} consists of 15,000 remote sensing images divided into 25 scene classes. We ask the model to classify each image into one of the 25 categories.
\item \textit{UC Merced}~\citep{UCMerced} is a 21 class land use image dataset. The images were manually extracted from large images from the USGS National Map Urban Area Imagery collection for various urban areas around the country. We ask the model to classify each image into one of the 21 categories.
\item \textit{FloodNet}~\citep{FloodNet} is a UAV imagery dataset captured after Hurricane Harvey, with visual question answering that challenges models to detect flooded roads and buildings and distinguish between natural and floodwater.
\item \textit{NWPU-Captions}~\citep{NWPU} includes 31,500 
images with annotated captions. The superiority of it lies in its wide coverage of complex scenes and the richness and variety of describing vocabularies. We evaluate models by taking an image and asking for the caption.
\end{itemize}

\begin{table}[!hb]
\centering
\resizebox{\columnwidth}{!}{%
\begin{tabular}{lllr}
\toprule
\textbf{Task}  & \textbf{Description}            & \textbf{Metric}   & \textbf{Test Num}  \\\midrule
\multicolumn{4}{l}{\hspace{-0.22cm}{\ul \textit{Biomedicine}}} \vspace{0.09cm} \\  
SLAKE$_{open}$     & VQA      & Recall   &  645 \\
SLAKE$_{clos.}$   & Binary classification   & Acc & 416 \\
PathVQA$_{open}$   & VQA      & Recall   & 3,357 \\
PathVQA$_{clos.}$ & Binary classification   & Acc & 3,362 \\
VQA-RAD$_{open}$   & VQA     & Recall   & 179  \\
VQA-RAD$_{clos.}$ & Binary classification   & Acc & 272 \\
PMC-VQA        & Multi-chioice QA        & Acc & 2,000  \\ \cmidrule{1-4}
\multicolumn{4}{l}{\hspace{-0.22cm}{\ul \textit{Food}}} \vspace{0.09cm} \\  
Recipe1M       & Recipe generation               & Rouge-L  & 1,000 \\
Nutrition5K    & Ingredient prediction           & Recall  & 507 \\
Food101        & Category classification             & Acc & 25,250 \\
FoodSeg103     & Multi-label classification & F1       & 2,135  \\ \cmidrule{1-4}
\multicolumn{4}{l}{\hspace{-0.22cm}{\ul \textit{Remote Sensing}}} \vspace{0.09cm} \\  
CLRS       & Scene classification               & Acc  & 15,000 \\
UC Merced    & Land-use classification      & Acc  & 21,000 \\
FloodNet        &  VQA           & Acc & 11,000 \\
NWPU     & Image captioning & Rouge-L       & 31,500  \\ \bottomrule
\end{tabular}
}
\caption{\textbf{Specifications of the Evaluated Domain-Specific Task Datasets.} }
\label{tab:tasks}
\end{table}

\begin{table*}[!htb]
\centering
\resizebox{\linewidth}{!}{%
\begin{tabular}{lll}
\toprule
\textbf{Task} & \textbf{Instruction}                                                                                    & \textbf{Response} \\ \midrule
\multicolumn{3}{l}{\hspace{-0.22cm}{\ul \textit{Biomedicine}}} \vspace{0.09cm} \\  
SLAKE       & {\texttt{\{question\}}}                                                                                                                                                                                                                                                                           & {\texttt{\{answer\}}}      \\ \cmidrule{1-3}
PathVQA     & {\texttt{\{question\}}}                                                                                                                                                                                                                                                                           & {\texttt{\{answer\}}}      \\ \cmidrule{1-3}
VQA-RAD     & {\texttt{\{question\}}}                                                                                                                                                                                                                                                                           & {\texttt{\{answer\}}}      \\ \cmidrule{1-3}
PMC-VQA     & \begin{tabular}[c]{@{}l@{}} {\texttt{Question:~\{question\}}}\\ {\texttt{The~choices~are:~\{options\}}}\end{tabular}                                                                                                                                                                                     & {\texttt{\{option\}}}      \\ \midrule
\multicolumn{3}{l}{\hspace{-0.22cm}{\ul \textit{Food}}} \vspace{0.09cm} \\  

Recipe1M    & {\texttt{\{question\}}}                                                                                                                                                                                                                                                                           & {\texttt{\{recipe\}}}     \\ \cmidrule{1-3}
Nutrition5K & {\texttt{What ingredients are used to make the dish in the image?}}                                                                                                                                                                                                                               & {\texttt{\{ingredients\}}} \\ \cmidrule{1-3}
Food101     & \begin{tabular}[c]{@{}l@{}}{\texttt{What type of food is shown in this image?}}\\ {\texttt{Choose one type from the following options:}}\\ {\texttt{\{food type options\}}}\end{tabular}                                                                                                                                & {\texttt{\{food type\}}}  \\ \cmidrule{1-3}
FoodSeg103  & \begin{tabular}[c]{@{}l@{}}{\texttt{Identify the food categories present in the image.}}\\ {\texttt{The available categories are: \{options\}}}\\ {\texttt{Please return a list of the selected food categories,}}\\{\texttt{formatted as a list of names like}} \\{\texttt{{[}candy, egg tart, french fries, chocolate{]}.}}\end{tabular} & {\texttt{\{categories\}}} \\  \midrule

\multicolumn{3}{l}{\hspace{-0.22cm}{\ul \textit{Remote Sensing}}} \vspace{0.09cm} \\  

CLRS    & \begin{tabular}[c]{@{}l@{}}{\texttt{What is the category of this remote sensing image?}}\\ {\texttt{Answer the question using a single word or phrase.}}\\  {\texttt{Reference categories include: \{scene options\}}}\end{tabular}                                                                                                                                                                                                                                                                           & {\texttt{\{scene category\}}}     \\ \cmidrule{1-3}
UC Merced    & \begin{tabular}[c]{@{}l@{}}{\texttt{What is the category of this remote sensing image?}}\\ {\texttt{Answer the question using a single word or phrase.}}\\  {\texttt{Reference categories include: \{land-use options\}}}\end{tabular}                                                                                                                                                                                                                                                                           & {\texttt{\{land-use category\}}}     \\ \cmidrule{1-3}
FloodNet       & {\texttt{\{question\}}}                                                                                                                                                                                                                                                                           & {\texttt{\{answer\}}}      \\ \cmidrule{1-3}
NWPU-Captions       &\begin{tabular}[c]{@{}l@{}} {\texttt{Please provide an one-sentence caption for the}}\\ {\texttt{provided remote sensing image in details.}}\end{tabular}                                                                                                                                                                                                                                                                          & {\texttt{\{caption\}}}      \\ \bottomrule
\end{tabular}
}
\caption{\textbf{Prompt Templates of the Evaluated Domain-Specific Task Datasets.} }
\label{tab:tasks_template}
\end{table*}

\begin{figure*}[!htb]
    \centering
    \includegraphics[width=\linewidth]{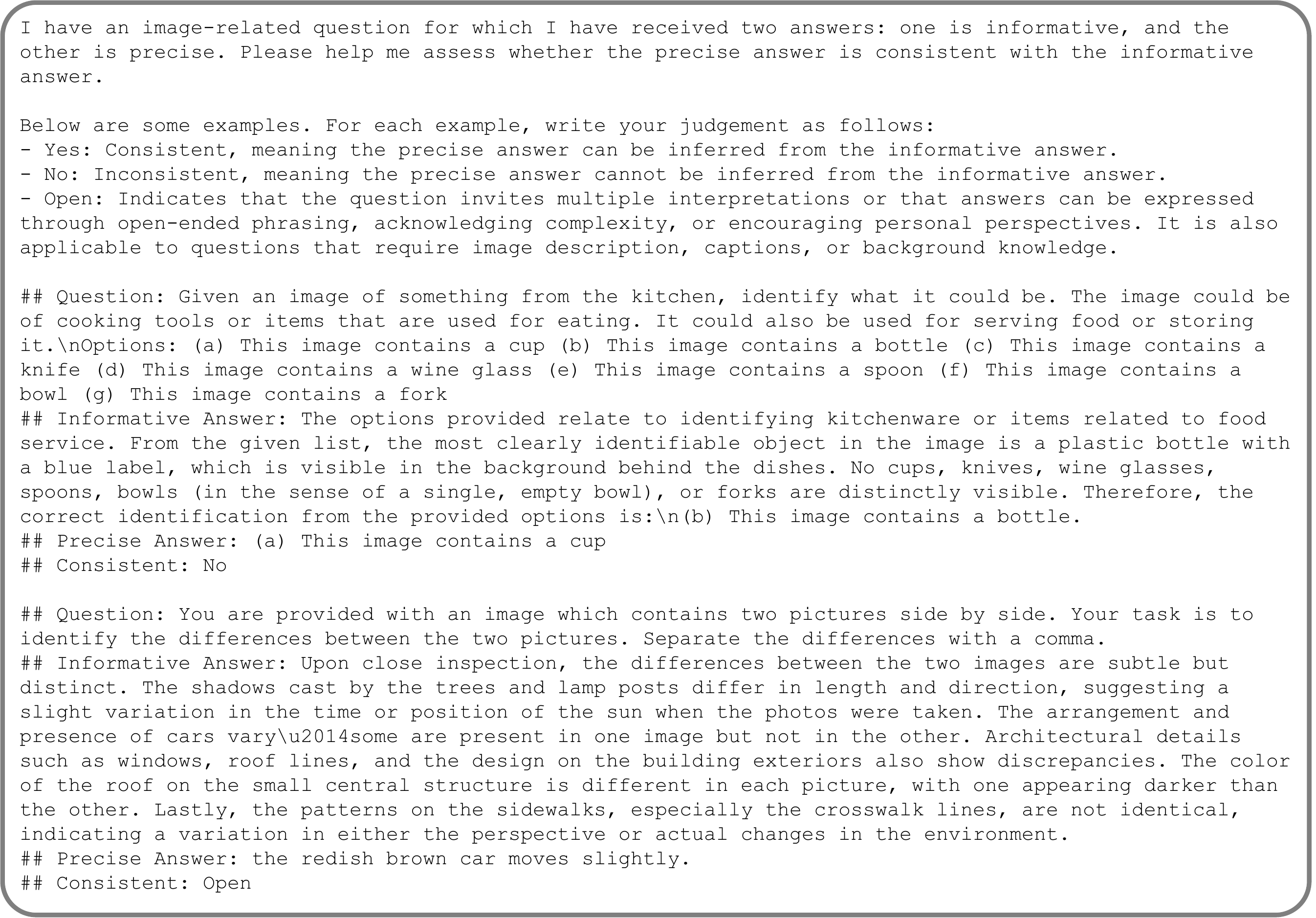}
    \caption{\textbf{Prompt Template for Consistency-Based Filter (Part 1)}, continued in Part 2.} 
\label{fig:consistency_prompt_1}
\end{figure*}

\begin{figure*}[!htb]
    \centering
    \includegraphics[width=\linewidth]{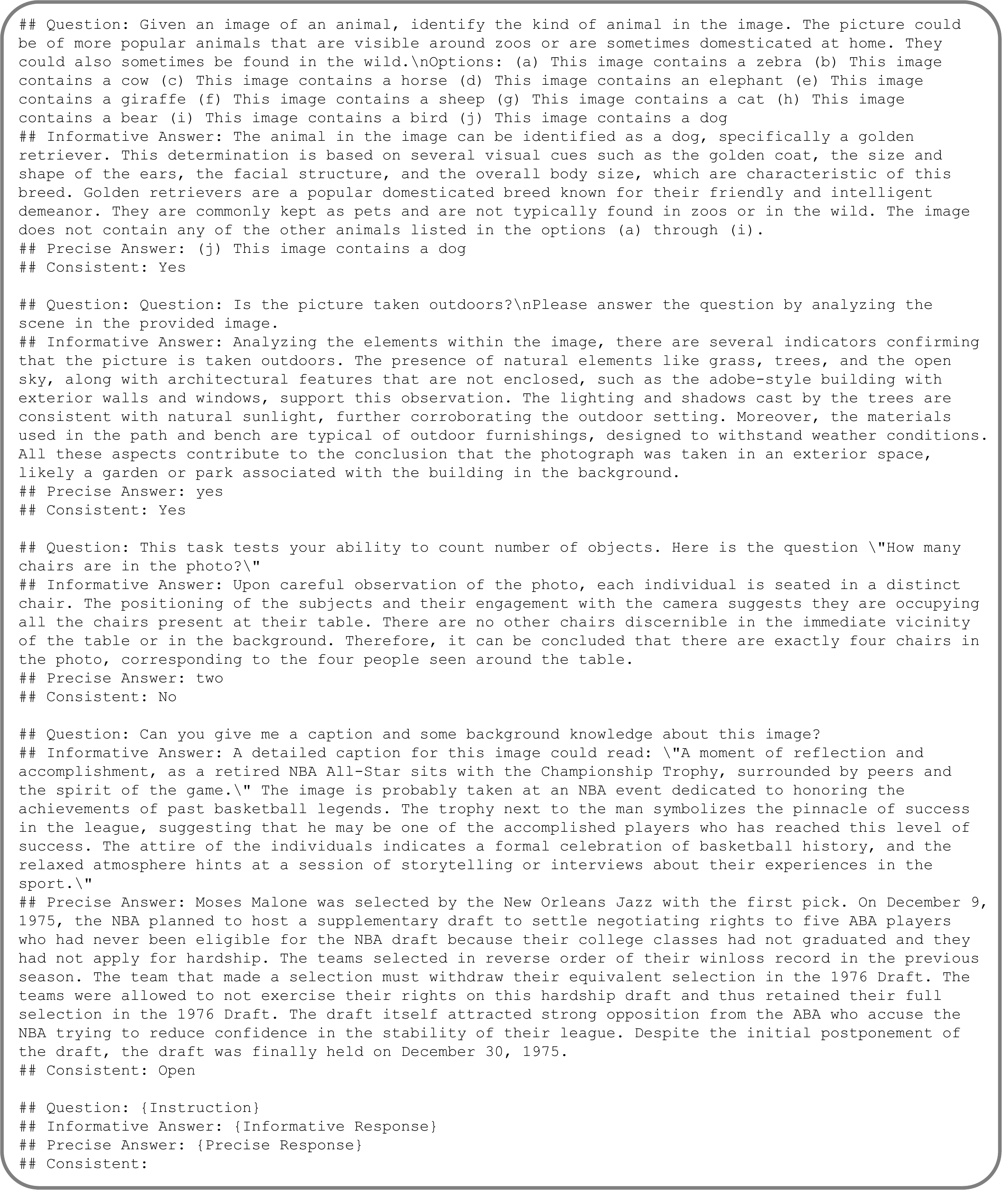}
    \caption{\textbf{Prompt Template for Consistency-Based Filter (Part 2).}} 
\label{fig:consistency_prompt_2}
\end{figure*}

\begin{table*}[!hb]
\centering
\resizebox{0.8\linewidth}{!}{\begin{tabular}{lccccccc}
\toprule
\textbf{\recipe}    & \textit{Train Pipeline} & {Instruction} & \textbf{Recipe} & \textbf{Nutrition} & \textbf{Food101} & \textbf{FoodSeg} & \textbf{\textsc{Average}}  \\\midrule
\multirow{4}{*}{LLaVA-v1.6-8B}    & \multirow{2}{*}{\textit{Two-Stage}} & Rule & 23.1     & 29.1      & 46.8    & 14.5    & 28.4 \\
                               &                            & Ours & 16.2     & 28.3      & 43.5    & 28.0    & 29.0 \\ \cmidrule(lr){2-8}
                               & \multirow{2}{*}{\textit{Single-Stage}}               & Rule & 21.8     & 36.7      & 63.9    & 13.9    & 34.1 \\
                               &                            & Ours & 24.8     & 36.1      & 65.3    & 42.0    & 42.0 \\ \cmidrule(lr){1-8}
\multirow{4}{*}{Qwen2-VL-2B}   & \multirow{2}{*}{\textit{Two-Stage}}                  & Rule & 24.1     & 24.5      & 68.8    & 7.7     & 31.3 \\
                               &                            & Ours & 16.5     & 43.0      & 69.5    & 23.9    & 38.2 \\ \cmidrule(lr){2-8}
                               & \multirow{2}{*}{\textit{Single-Stage}}               & Rule & 19.3     & 37.1      & 64.7    & 6.6     & 31.9 \\
                               &                            & Ours & 24.0     & 41.2      & 72.0    & 23.9    & 40.3 \\\cmidrule(lr){1-8}
\multirow{4}{*}{Llama-3.2-11B} & \multirow{2}{*}{\textit{Two-Stage}}                  & Rule & 25.7     & 26.2      & 82.1    & 16.7    & 37.7 \\
                               &                            & Ours & 17.8     & 38.0      & 74.6    & 33.2    & 40.9 \\ \cmidrule(lr){2-8}
                               & \multirow{2}{*}{\textit{Single-Stage}}               & Rule & 21.4     & 32.2      & 75.8    & 16.9    & 36.6 \\
                               &                            & Ours & 26.1     & 41.0      & 82.2    & 42.0    & 47.8 \\ \bottomrule
\end{tabular}
}
\vspace{15pt}

\resizebox{\linewidth}{!}{\begin{tabular}{lcccccccccc}
\toprule
\multirow{2}{*}{\textbf{\pmcraw}} & \multirow{2}{*}{\textit{Train Pipeline}} & \multicolumn{1}{c}{\multirow{2}{*}{Instruction}} & \multicolumn{2}{c}{\textbf{SLAKE}} & \multicolumn{2}{c}{\textbf{PathVQA}} & \multicolumn{2}{c}{\textbf{VQA-RAD}} & \multicolumn{1}{c}{\multirow{2}{*}{\textbf{PMC-VQA}}} & \multicolumn{1}{c}{\multirow{2}{*}{\textbf{\textsc{Average}}}} \\ \cmidrule(lr){4-5} \cmidrule(lr){6-7} \cmidrule(lr){8-9}
                                        &                                 & \multicolumn{1}{c}{}                             & OPEN       & CLOSED       & OPEN     & CLOSED           & OPEN        & CLOSED        & \multicolumn{1}{c}{}                         & \multicolumn{1}{c}{}                         \\   \midrule
\multirow{4}{*}{LLaVA-v1.6-8B}          & \multirow{2}{*}{\textit{Two-Stage}}      & GPT-4                                            & 43.4       & 50.2         & 10.1     & 59.2             & 35.0        & 62.5          & 37.1                                         & 42.5                                         \\
                                        &                                 & Ours                                             & 56.2       & 71.4         & 17.2     & {74.5}    & 50.6        & 79.0          & {40.4}                                & 55.6                                         \\ \cmidrule(lr){2-11}
                                        & \multirow{2}{*}{\textit{Single-Stage}}   & GPT-4                                            & 44.2       & 59.1         & 11.6     & 62.2             & 38.5        & 67.3          & 39.9                                         & 46.1                                         \\
                                        &                                 & Ours                                             & 56.8       & 76.4         & 19.7     & 79.3             & 51.0        & 80.5          & 44.3                                         & 58.3                                         \\ \cmidrule(lr){1-11}
\multirow{4}{*}{Qwen2-VL-2B}            & \multirow{2}{*}{\textit{Two-Stage}}      & GPT-4                                            & 43.4       & 55.5         & 11.8     & 60.1             & 37.1        & 58.8          & 41.2                                         & 44.0                                         \\
                                        &                                 & Ours                                             & 55.2       & 74.5         & 18.4     & 68.4             & 48.8        & 79.8          & 43.8                                         & 55.5                                         \\ \cmidrule(lr){2-11}
                                        & \multirow{2}{*}{\textit{Single-Stage}}   & GPT-4                                            & 43.6       & 59.6         & 13.2     & 47.4             & 37.3        & 57.0          & 31.2                                         & 41.3                                         \\
                                        &                                 & Ours                                             & 53.2       & 75.2         & 20.1     & 63.8             & 49.8        & 74.6          & 43.5                                         & 54.3                                         \\ \cmidrule(lr){1-11}
\multirow{4}{*}{Llama-3.2-11B}          & \multirow{2}{*}{\textit{Two-Stage}}      & GPT-4                                            & 47.6       & 58.7         & 14.6     & 69.5             & 38.0        & 69.1          & 47.5                                         & 49.3                                         \\
                                        &                                 & Ours                                             & 60.0       & 75.7         & 22.1     & 76.8             & 51.4        & 80.5          & 47.9                                         & 59.2                                         \\ \cmidrule(lr){2-11}
                                        & \multirow{2}{*}{\textit{Single-Stage}}   & GPT-4                                            & 46.8       & 56.5         & 16.0     & 69.9             & 41.9        & 65.4          & 45.3                                         & 48.8                                         \\ 
                                        &                                 & Ours                                             & 56.7       & 77.6         & 22.2     & 87.3             & 55.0        & 76.1          & 49.9                                         & 60.7         \\ \bottomrule                            
\end{tabular}
}

\vspace{15pt}

\resizebox{\linewidth}{!}{\begin{tabular}{lcccccccccc}
\toprule
\multirow{2}{*}{\textbf{\pmcrefine}} & \multirow{2}{*}{\textit{Train Pipeline}} & \multicolumn{1}{c}{\multirow{2}{*}{Instruction}} & \multicolumn{2}{c}{\textbf{SLAKE}} & \multicolumn{2}{c}{\textbf{PathVQA}} & \multicolumn{2}{c}{\textbf{VQA-RAD}} & \multicolumn{1}{c}{\multirow{2}{*}{\textbf{PMC-VQA}}} & \multicolumn{1}{c}{\multirow{2}{*}{\textbf{\textsc{Average}}}} \\ \cmidrule(lr){4-5} \cmidrule(lr){6-7} \cmidrule(lr){8-9}
                                        &                                 & \multicolumn{1}{c}{}                             & OPEN       & CLOSED       & OPEN     & CLOSED           & OPEN        & CLOSED        & \multicolumn{1}{c}{}                         & \multicolumn{1}{c}{}                         \\   \midrule
\multirow{4}{*}{LLaVA-v1.6-8B}             & \multirow{2}{*}{\textit{Two-Stage}}      & GPT-4V                                           & 50.0       & 68.3         & 17.0        & 67.5          & 43.3        & 67.3          & 40.4                                         & 50.5                                         \\
                                           &                                 & Ours                                             & 54.8       & 73.1         & 19.3        & 79.7          & 55.6        & 82.7          & 45.1                                         & 58.6                                         \\ \cmidrule(lr){2-11}
                                           & \multirow{2}{*}{\textit{Single-Stage}}   & GPT-4V                                           & 52.3       & 76.2         & 20.1        & 73.3          & 47.0        & 76.5          & 43.1                                         & 55.5                                         \\
                                           &                                 & Ours                  & 58.0        & 73.3          & 22.9        & 78.6                            & 59.8       & 81.3                  & 47.9                                         & 60.3                                         \\ \cmidrule(lr){1-11}
\multirow{4}{*}{Qwen2-VL-2B}               & \multirow{2}{*}{\textit{Two-Stage}}      & GPT-4V                                           & 45.2       & 63.2         & 18.2        & 64.7          & 41.3        & 67.3          & 43.2                                         & 49.0                                         \\
                                           &                                 & Ours                                             & 60.8       & 76.9         & 21.4        & 75.0          & 55.0        & 82.7          & 44.7                                         & 59.5                                         \\ \cmidrule(lr){2-11}
                                           & \multirow{2}{*}{\textit{Single-Stage}}   & GPT-4V                                           & 51.4       & 66.1         & 18.9        & 61.4          & 45.1        & 73.2          & 45.1                                         & 51.6                                         \\
                                           &                                 & Ours                                             & 60.2       & 75.0         & 20.6        & 53.6          & 58.0        & 76.1          & 46.5                                         & 55.7                                         \\\cmidrule(lr){1-11}
\multirow{4}{*}{Llama-3.2-11B}             & \multirow{2}{*}{\textit{Two-Stage}}      & GPT-4V                                           & 49.1       & 74.3         & 19.3        & 70.9          & 46.2        & 73.9          & 47.1                                         & 54.4                                         \\
                                           &                                 & Ours                                             & 58.5       & 76.4         & 27.0        & 73.2          & 58.3        & 77.6          & 51.3                                         & 60.3                                         \\\cmidrule(lr){2-11}
                                           & \multirow{2}{*}{\textit{Single-Stage}}   & {GPT-4V}                                  & 47.1       & 72.6         & 19.5        & 70.7          & 45.9        & 73.9          & 46.5                                         & 53.7                                         \\
                                           &                                 & {Ours}                                    & 59.5       & 76.4         & 24.3        & 84.9          & 57.4        & 79.8          & 51.9                                         & 62.0  \\                                      
\bottomrule                            
\end{tabular}
}
\caption{\textbf{Domain-Specific Task Performance of MLLMs after Post-Training} with different synthetic data and training pipelines. The image-caption sources are \recipe, \pmcraw~and~\pmcrefine, respectively.
In most cases using our synthetic data, we find that single-stage training outperforms two-stage training on domain-specific tasks, particularly evident in the \textbf{Recipe} generation results for the food domain. Recall that in the two-stage training approach for the food domain, the model first trains on recipe captions and then on our synthetic tasks. We examine the task performance of LLaVA-v1.6-8B on \textbf{Recipe} generation and observe that the model achieves a score of 25.3 after the first stage on recipe captions. However, this score drastically decreases to 16.2 after the second stage. From this, we infer that the two-stage approach causes the model to catastrophically forget the task/knowledge learned in the first stage when transitioning to the second stage~\citep{catastrophic-forget}, leading to poorer performance after completing the second-stage training.}
\label{tab:detailed Comparison of Synthetic Tasks and Training Pipeline}
\end{table*}
\clearpage

\begin{figure*}[!htb]
    \centering
    \includegraphics[width=\linewidth]{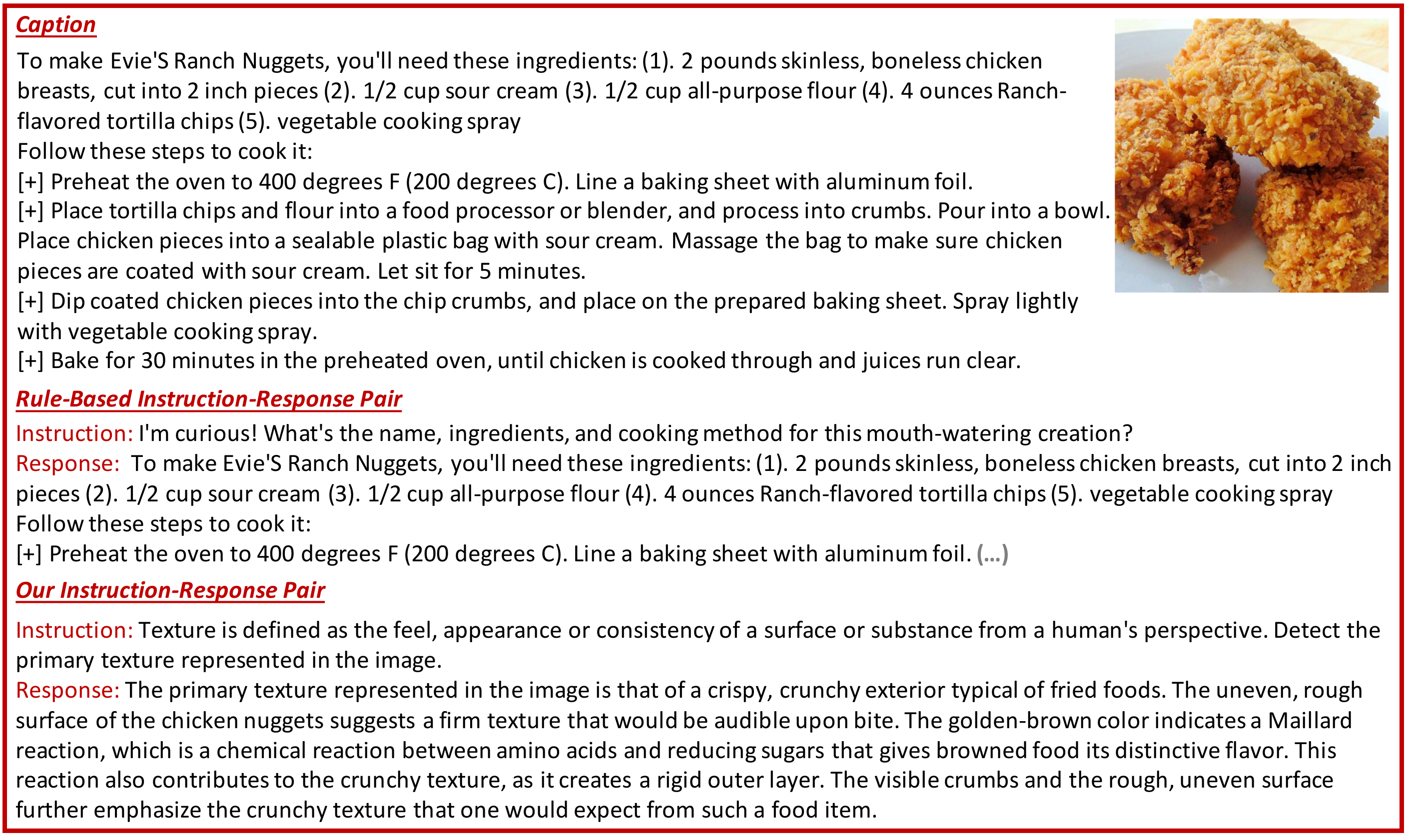}
    
    \vspace{10pt}
    
    \includegraphics[width=\linewidth]{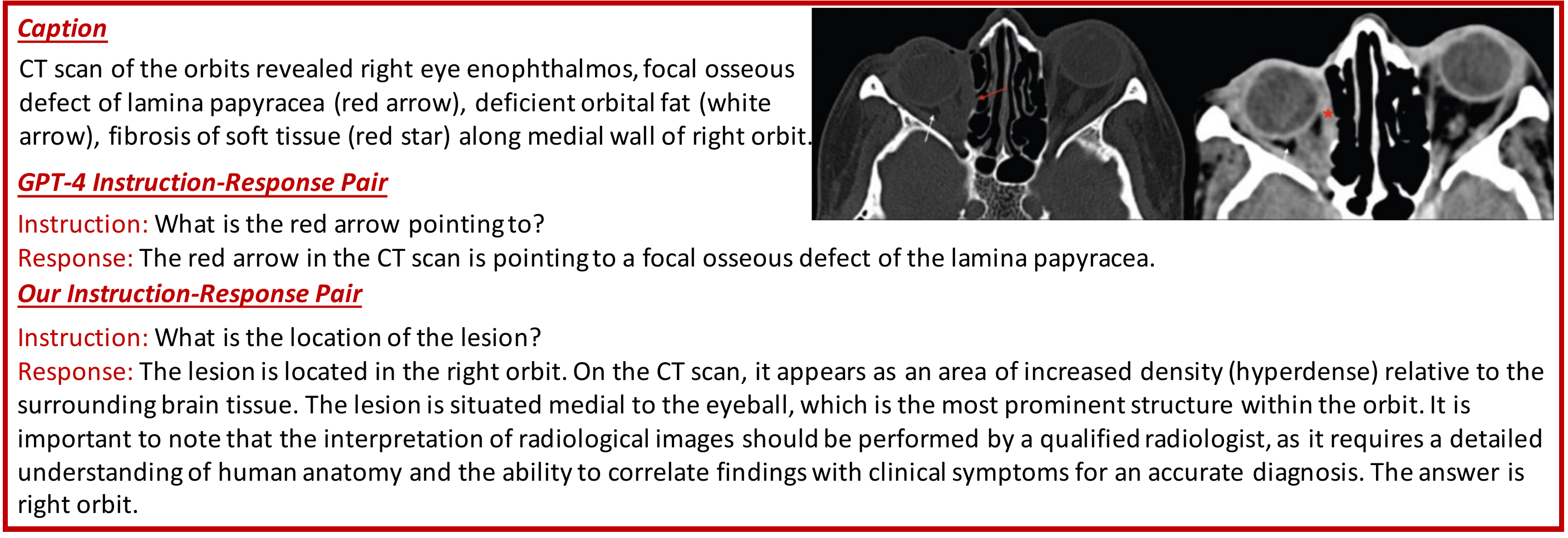}
    \caption{\textbf{Cases of Instruction-Response Pairs (Part 1)} synthesized by our method, manual rules, GPT-4, and GPT-4V, the image-caption sources for the cases are \recipe, \pmcraw~and \pmcrefine, respectively. Continued in Part 2. In the first case, the rule-based task simply transforms the recipe caption, ignoring the image content. In contrast, our task involves analyzing the food's state in the image and applying food-related knowledge to infer its texture, demonstrating a higher level of {\ul domain knowledge utilization}. In the second case, the GPT-4 generated task straightforwardly asks about the pointing of the red arrow, while ours requires a detailed analysis and inference, showing greater {\ul task complexity}.} 
\label{fig:more_case_1}
\end{figure*}

\begin{figure*}[!htb]
    \centering
    \includegraphics[width=\linewidth]{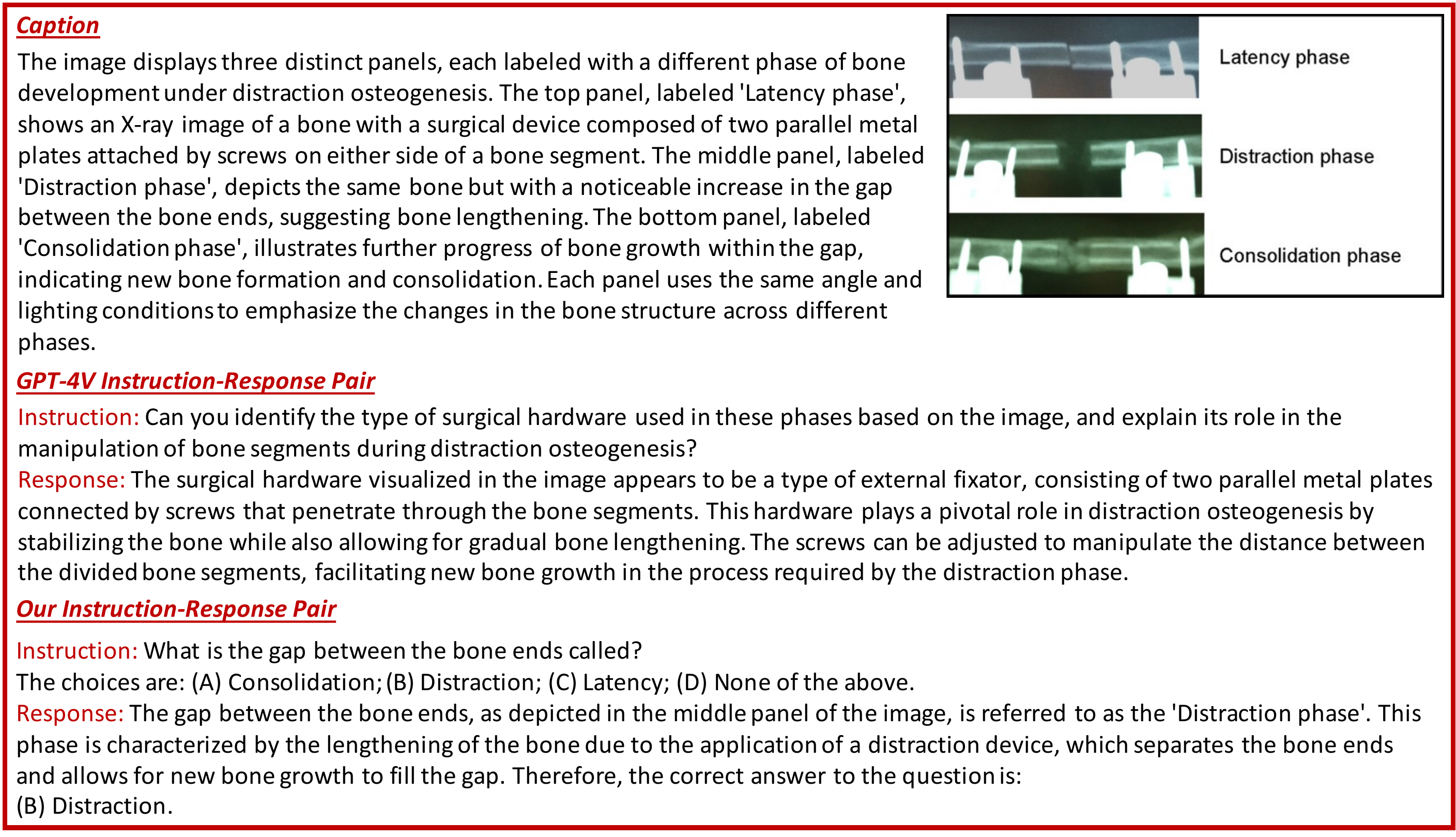}
    \caption{\textbf{Cases of Instruction-Response Pairs (Part 2)} synthesized by our method, manual rules, GPT-4, and GPT-4V, the image-caption sources for the cases are \recipe, \pmcraw~and \pmcrefine, respectively. In this case, our task stands out as a multiple-choice question, showcasing {\ul task diversity}.} 
\label{fig:more_case_2}
\end{figure*}

\end{document}